\documentclass[8pt]{extarticle}

\usepackage{spconf,amsmath,epsfig}
\usepackage{amssymb}
\usepackage{booktabs}
\usepackage{subfigure}
\usepackage{multirow}
\usepackage{hyperref}
\usepackage{color}
\usepackage{floatrow}
\usepackage[figuresright]{rotating}

\newfloatcommand{capbtabbox}{table}[][\FBwidth]

\let\OLDthebibliography\thebibliography
\renewcommand\thebibliography[1]{
  \OLDthebibliography{#1}
  \setlength{\parskip}{0pt}
  \setlength{\itemsep}{0pt plus 0.3ex}
}

\begin{document}\sloppy

\def\x{{\mathbf x}}
\def\L{{\cal L}}

\title{Deep Transformers Thirst for Comprehensive-Frequency Data}
%
\name{Rui Xia,  Chao Xue , Boyu Deng , Fang Wang , Jingchao Wang$^\star$ \thanks{$^\star$ Corresponding author}}
\address{
Academy of Military Science of the People's Liberation Army, Beijing, China\\
$xiarui414@foxmail.com, \left\{wangjc.2000, xuec11\right\}@tsinghua.org.cn, kdydby2014@sina.com$}

\maketitle

\begin{abstract}
Current researches indicate that inductive bias (IB) can improve Vision Transformer (ViT) performance. However, they introduce a pyramid structure concurrently to counteract the incremental FLOPs and parameters caused by introducing IB. This structure destroys the unification of computer vision and natural language processing (NLP) and complicates the model. We study an NLP model called LSRA \cite{Wu2020Lite}, which introduces IB with a pyramid-free structure. We analyze why it outperforms ViT, discovering that introducing IB increases the share of high-frequency data in each layer, giving "attention" to more information. As a result, the heads notice more diverse information, showing better performance. To further explore the potential of transformers, we propose EIT, which Efficiently introduces IB to ViT with a novel decreasing convolutional structure under a pyramid-free structure. EIT achieves competitive performance with the state-of-the-art (SOTA) methods on ImageNet-1K and achieves SOTA performance over the same scale models which have the pyramid-free structure.
\end{abstract}
\renewcommand{\thefootnote}{}
\footnotetext{Supported by the National Natural Science Foundation of China (Grant No. 62022093 and No. 62101587).}
\footnotetext{Code:https://github.com/MrHaiPi/EIT}

\section{Introduction}
\begin{figure}[htbp]
\centering
\includegraphics[width=0.8\textwidth]{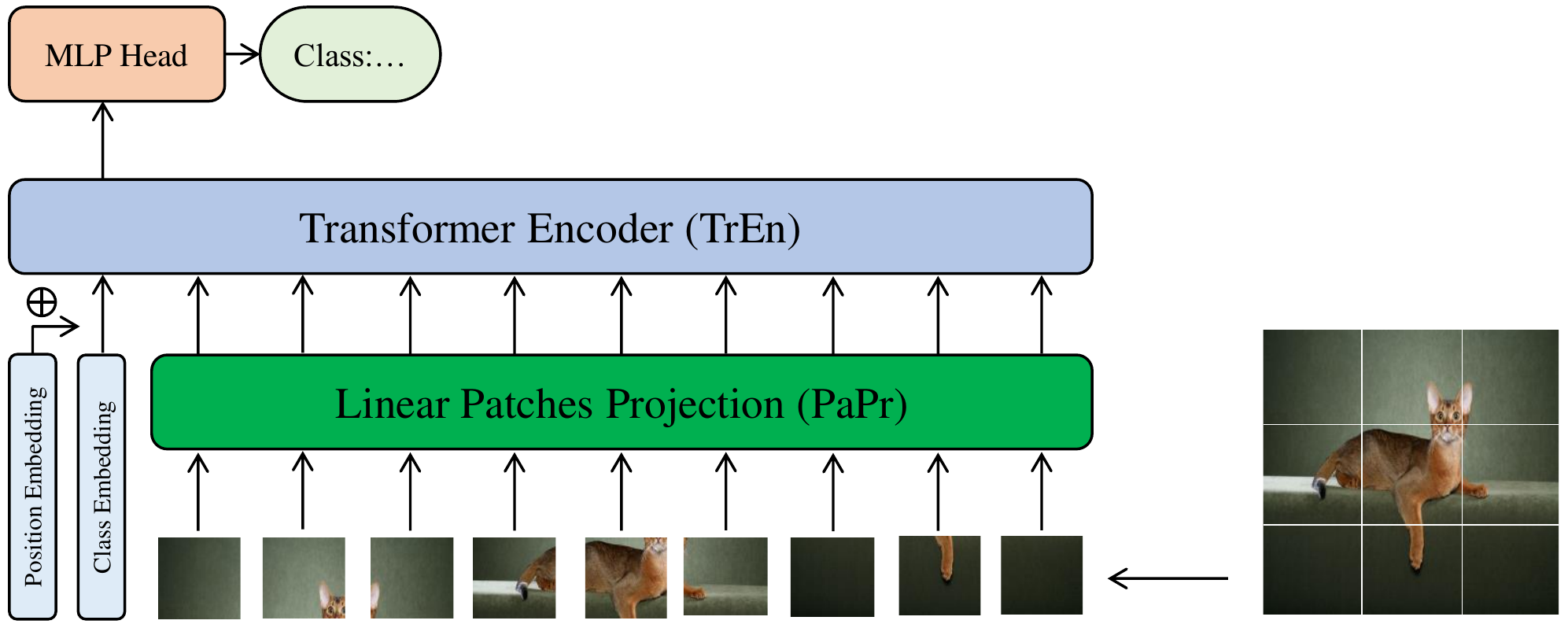}
\caption{The brief structure of ViT.}
\label{fig:ViTBriefStru}
\vspace{-0.35cm}
\end{figure}

In recent years, Transformer \cite{Vaswani2017Attention} has swept the field of natural language processing (NLP) due to its superior performance. The fields of the computer vision (CV) \cite{Dosovitskiy2021An,Wu2020Visual} and even the multi-agent reinforcement learning (MARL) \cite{RN173,RN174} are also gradually infiltrated by this famous technology. Vision Transformer (ViT) \cite{Dosovitskiy2021An} is the first CV model which is completely based on the Transformer architecture and achieves better performance on ultra-large-scale image classification task than the convolutional neural networks (CNN). ViT is also the first model trying to break the barrier of unifying the CV and NLP with the same backbone and achieved a breakthrough. Such a model is beneficial for the future research on the MultiModal networks \cite{ngiam2011multimodal}. Specifically, ViT divides images into several non-overlapping patches to correspond the $tokens$ in NLP, and uses the full-Transformer architecture to model the patches and complete the image classification task. The brief structure of ViT is shown in Fig.\ref{fig:ViTBriefStru}.

\begin{figure}[htbp]
\centering
\includegraphics[width=0.7\textwidth]{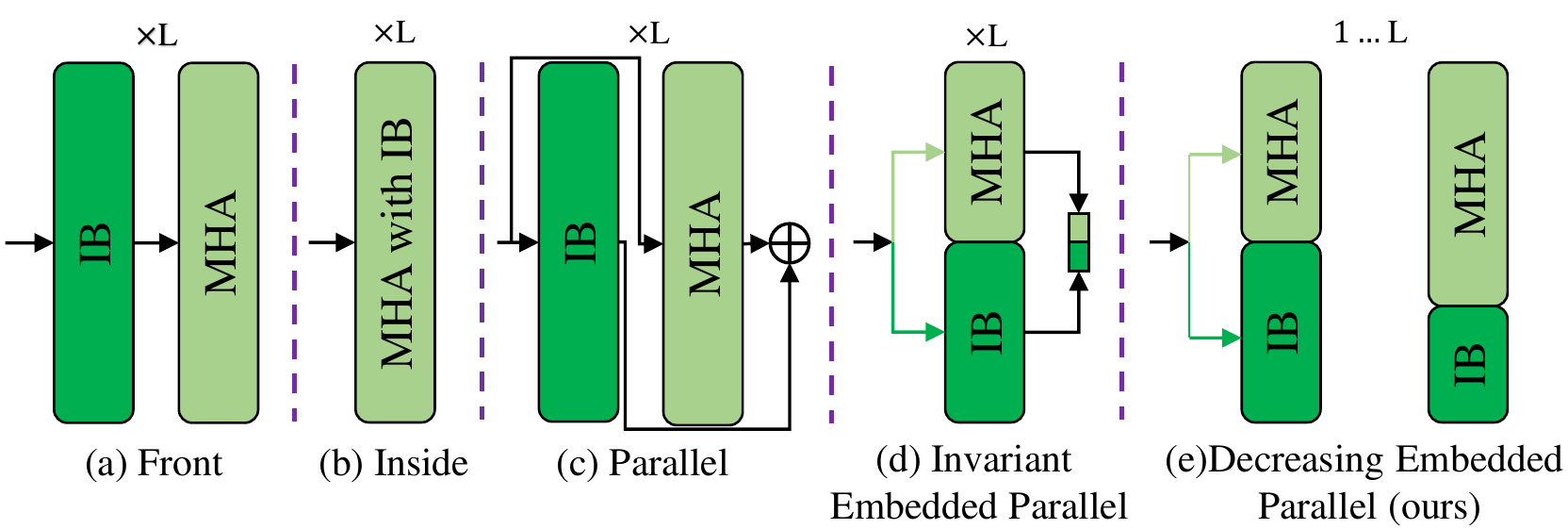}
\vspace{-0.35cm}
\caption{The ways of introducing IB. (a) Data is pre-processed by IB structure \cite{Han2021Transformer,liu2021swin,Yuan2021Tokens}. (b) MHA with built-in IB \cite{Wu2021CvT}. (c) MHA handles all data together with the module with IB structure \cite{wang2021evolving}. (d) The MHA and the IB structure each process a invariant portion of the data \cite{Wu2020Lite}. (e) IB structure in the previous layer process less data than those in the next layer.}
\label{fig:1}
\vspace{-0.25cm}
\end{figure}

The original ViT relies on the ultra-large-scale datasets to perform better than the same scale CNN \cite{Dosovitskiy2021An}. The strong data augmentation can free ViT from the dependence on the ultra-large-scale datasets \cite{Touvron2021Training}. However, many studies have shown that ViT trained with strong data augmentation is still suboptimal \cite{Han2021Transformer,liu2021swin,Yuan2021Tokens,Wu2021CvT,wang2021evolving,Xiao2021Early}. One reason is that Transformers lack some of the properties similar to inductive biases (IB) inherent to CNN, such as locality \cite{Dosovitskiy2021An}. Many studies have been explored on introducing the IB into ViT, such as the locality-based methods like TNT \cite{Han2021Transformer} and T2T \cite{Yuan2021Tokens}, or CNN based methods like EA \cite{wang2021evolving}, CvT \cite{Wu2021CvT}, Swin \cite{liu2021swin} and ViT$_C$ \cite{Xiao2021Early}. In summary, the recent methods of introducing IB can be summarized in three categories, as shown in the first three items of Fig.\ref{fig:1}. The aforementioned researches have experimentally demonstrated that the introducing IB can improve the performance of ViT. Since the introducing IB brings the redundant structures, a few of them suffer from a significant increase in parameters and FLOPs \cite{Han2021Transformer,Yuan2021Tokens,wang2021evolving}. To solve this problem, quite a few of them further introduce the pyramid structure \cite{liu2021swin,Wu2021CvT,Xiao2021Early}. However, the pyramid structure destroys the structure of Transformer and departs from the original intention of ViT, which intends to unify CV and NLP by a same structure. Moreover, the pyramid structure complicates the model, which is not conducive to further optimization.

In this work, to ensure the unification of CV and NLP and keep the model brevity while introducing IB, we study LSRA \cite{Wu2020Lite}, a model of NLP. Its brief structure is shown in Fig.\ref{fig:1}-(d), which divides the data stream along the channel dimension to Multi-Head Attention (MHA) and IB, achieving introducting IB without increasing FLOPs and parameters. We analyze why it outperforms ViT finding that the lack of IB in ViT is manifested explicitly by the small diversity of head-attention distances (Head Diversity) in the deep layers, which causes a degradation in the performance. We qualitatively analyze the reasons for the small Head Diversity of ViT's deep layers and explain how the LSRA introduces IB into ViT from the perspective of a filter. As a high-frequency filter, the IB improves the deep layers' share of high-frequency data, giving 'attention' more comprehensive information, thus boosting Head Diversity. However, we find LSRA suffers from the inefficient introducing IB, resulting in small Head Diversity and performance limitations.

Based on this, we propose a model called EIT, which can \textbf{E}fficiently introduce the \textbf{I}B to the Vi\textbf{T} under pyramid-free structures and with fewer parameters and FLOPs, as shown in Fig.\ref{fig:1}-(e). We make the following two changes to get the Decreasing Embedded
Parallel Structure (Decreasing Structure): 1) the linear Patches Projection (PaPr) of ViT is replaced by a convolutional layer plus a maximum pooling layer, which is called EIT$^P$; 2) a decreasing convolutional structure is introduced to the MHA of Transformer Encoder (TrEn), by which the embedding's different channel dimensions are processed, respectively. We validated the performance of EIT on four small-scale datasets and a large-scale dataset showing that the EIT outperforms the similar ViT-like methods (ViTs). Moreover, the proposed Decreasing Structure is generally applicable to Transformer-based architectures and further impacts a broader range of applications. We summarize the contributions as follows:
\begin{enumerate}
    \item To the best of our knowledge, we first find and explain that introducing IB can increase the share of high-frequency data in each layer, giving "attention" more comprehensive data. Comprehensive data improves the diversity of head-attention distances in Transformers, which causes performance improvement.
    \item We propose a novel Decreasing Structure based on the principle that MHA and IB are low-pass filters and high-pass filters. Such a structure can efficiently introduce IB to ViT with fewer parameters and FLOPs with a pyramid-free structure, ensuring the unification of CV and NLP.
    \item We conduct comprehensive experiments that show EIT achieves promising and competitive results than the similar representative state-of-the-art methods currently available. In particular, EIT achieves state-of-the-art performance over the other same scale models which have a pyramid-free structure.
\end{enumerate}

\section{Related Works}

\subsection{Vision Transformer}
Although there were many Transformer-based models in the CV field, ViT \cite{Dosovitskiy2021An} was the first model based entirely on Transformer and tried to unify the CV and NLP with the same network structure. In its implementation, ViT first split an image into non-overlapping patches, then mapped the patches into patches embedding by a linear mapping layer. Finally, it classified the images by connecting multiple standard TrEn.
However, the ViT, which is trained with ultra-large-scale datasets (e.g., ImageNet-21k and JFT-300M) \cite{Dosovitskiy2021An} and strong data augmentation (e.g., MixUp \cite{zhang2017mixup}, CutMix \cite{yun2019cutmix} and Erasing \cite{zhong2020random}), is still suboptimal, with the reason of IB lacking \cite{Han2021Transformer,liu2021swin,Yuan2021Tokens,Wu2021CvT,wang2021evolving,Xiao2021Early}. In this paper, we study how to efficiently introduce IB to ViT.

\subsection{Introduce Inductive Biases to ViT}
Many methods have been proposed to introduce IB to ViT. For example, Transformer-in-Transformer (TNT) \cite{Han2021Transformer} introduced the Transformer module inside patches to model a more detailed pixel-level representation. Tokens-to-Token (T2T) \cite{Yuan2021Tokens} stitched together neighbouring embedding to form a new embedding in the original location to change ViT's PaPr. Such an operation could preemptively improve the similarity of neighbouring, which in turn introduced IB. CvT \cite{Wu2021CvT} introduced IB to both PaPr and TrEn by convolution operation with a hierarchical structure. Specifically, CvT changed the original linear mapping of ViT to a convolutional mapping in both PaPr and TrEn. The stride of convolution was smaller than the kernel size. ViT$_C$ \cite{Xiao2021Early} changed the PaPr with Convolutional Stem to help transformers see better. Besides, in NLP, Long-Short Range Attention (LSRA) \cite{Wu2020Lite} introduced Lite Transformer Block (LTB) to TrEn, which divided half of the data to be processed by MHA along the channel dimension to the convolutional layer. Summarizing the above works, we can find two phases to introduce the IB to ViT. One is the PaPr phase, and the other is the TrEn phase. Table \ref{tab:1} summarizes the contributions of the above works.

\begin{table}[htbp]
\fontsize{6}{8}\selectfont
\centering
\vspace{-0.3cm}
\caption{Representative works of introducing IB.}\label{tab:1}
\setlength{\tabcolsep}{0.5mm}{
\begin{tabular}{|c|c|c|c|c|}
\hline
Method    &\begin{tabular}[c]{@{}c@{}} Position\\ Encoding \end{tabular}& \begin{tabular}[c]{@{}c@{}}Introduce IB \\to PaPr phase\end{tabular} & \begin{tabular}[c]{@{}c@{}}Introduce IB \\to TrEn phase\end{tabular} & Pyramid\\
\hline
LSRA \cite{Wu2020Lite}  & Cosine                      & None                                                                                               & \begin{tabular}[c]{@{}c@{}}Lite Transformer\\ Block (LTB)\end{tabular} & $\times$ \\
ViT \cite{Dosovitskiy2021An}    & Trainable                      & None                                                                                               & None &  $\times$                                                                                                      \\

T2T \cite{Yuan2021Tokens}   & Trainable                      & Concatenate                                                                                        & None      &  $\times$                                                                                                  \\
TNT \cite{Han2021Transformer}   & Trainable                      & Patch + Pixel                                                                                        & Patch + Pixel  &  $\times$                                                                                                       \\
CvT \cite{Wu2021CvT}   & None                       & \begin{tabular}[c]{@{}c@{}c@{}}Convolutional \\Embedding (CvT$^P$)\end{tabular}                           & \begin{tabular}[c]{@{}c@{}}Convolutional \\Mapping QKV (CvT$^T$)\end{tabular}    &$\checkmark$  \\
ViT$_C$ \cite{Xiao2021Early}    & Trainable                      & Convolutional Stem (CoSt)                                                                                              & None  &$\checkmark$                                                                                                      \\
EIT(ours) & None                       & \begin{tabular}[c]{@{}c@{}}Convolutional \\Embedding and Maxpool (EIT$^P$)\end{tabular}                 & \begin{tabular}[c]{@{}c@{}}Embedded Parallel \\Decreasing Convolution (EIT$^T$)\end{tabular}                &   $\times$  \\
\hline
\end{tabular}
}
\vspace{-0.3cm}
\end{table}

However, the above works suffer from the inefficient introducing IB or the inability to reconcile the contradiction between pyramid-free structures and fewer FLOPs/parameters. In contrast to these concurrent works, our method ensures both the efficiency of IB introduction and the pyramid-free structures without increasing the FLOPs/parameters.

\begin{figure}
\centering
\begin{minipage}[b]{.47\linewidth}
\centering
\vspace{-0.2cm}
\subfigure{\includegraphics[width=1\textwidth]{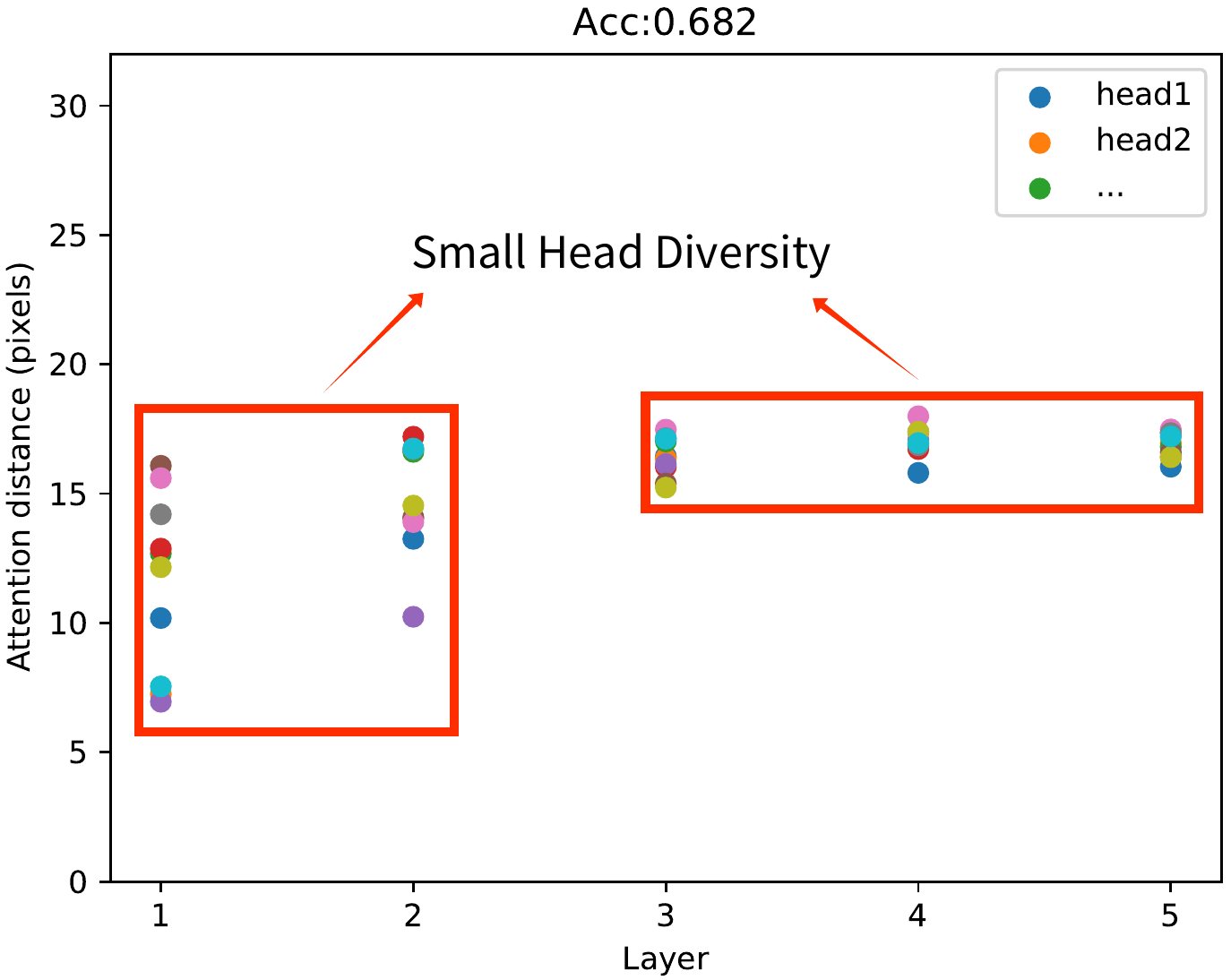}}
\vspace{-0.4cm}
\tiny{\center{~~~~~ViT \cite{Dosovitskiy2021An}.}}
\vspace{-0.4cm}
\end{minipage}\hspace{0.25cm}
\begin{minipage}[b]{.47\linewidth}
\vspace{-0.2cm}
\centering
\subfigure{\includegraphics[width=1\textwidth]{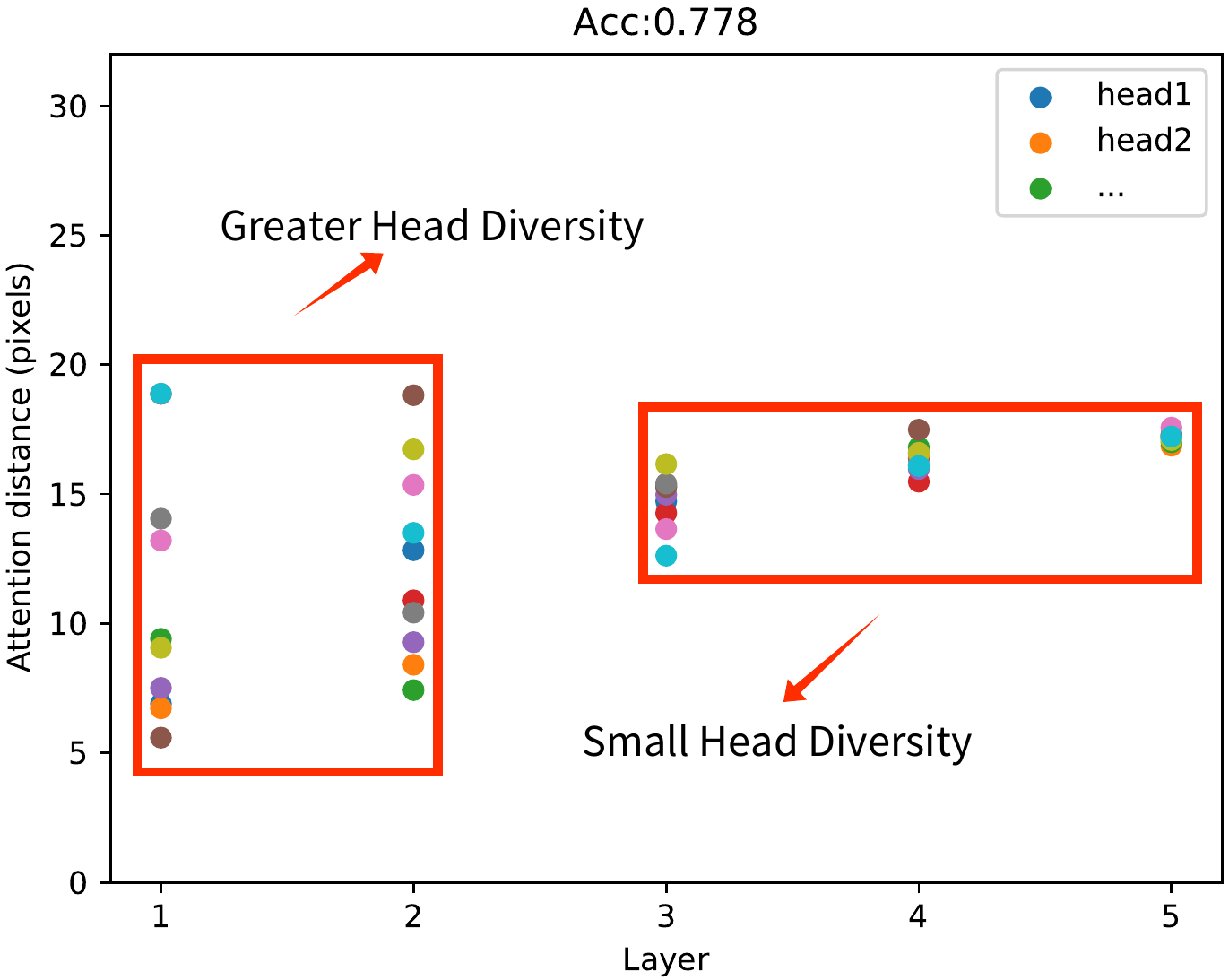}}
\vspace{-0.4cm}
\tiny{\center{LSRA \cite{Wu2020Lite}.}}
\vspace{-0.4cm}
\end{minipage}
\caption{The distribution of the head-attention distances for each layer. The models consist of five TrEns, and each TrEn contains ten heads. The above results are from training and testing on the Cifar10 \cite{Krizhevsky2009Learning}. Attention distance is computed for 2000 example images from Cifar10 by summing the distance between the selected query pixel and all other pixels, weighted by the attention weight \cite{Dosovitskiy2021An}. The variance of the attention distances at each layer approximates the Head Diversity. The figure shows that the Head Diversity of layers 3, 4, and 5 of LSRA is basically the same as ViT. The LSRA's Head Diversity of the first two layers are larger than ViT, thus performs better.}
\label{fig:2}

\end{figure}

\section{Our Approach}
\subsection{Motivation}
\begin{figure}[htb]
\centering
\includegraphics[width=0.5\textwidth]{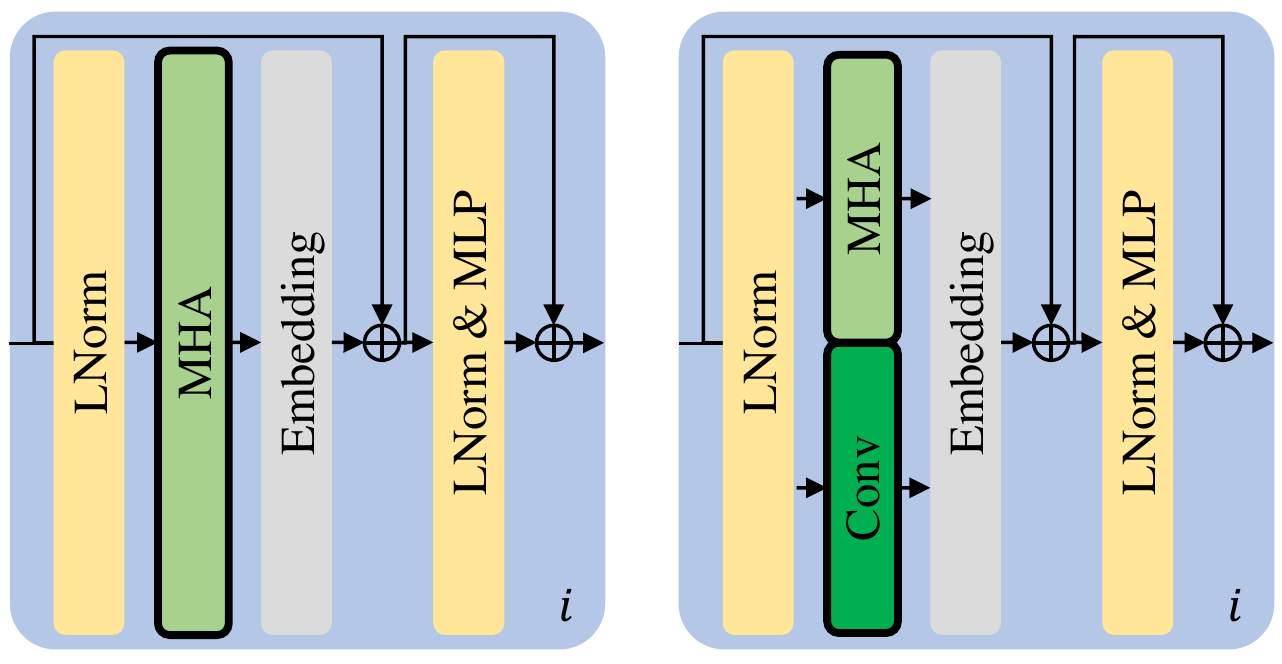}
\caption{The main structure of ViT (left) and LSRA (right).}
\label{fig:MainStruofViTandLSRA}
\vspace{-0.6cm}
\end{figure}

We determine that LSRA \cite{Wu2020Lite}, a model of NLP as shown in Fig.\ref{fig:MainStruofViTandLSRA} (right), can also performer better than ViT \cite{Dosovitskiy2021An} for image classification. When we investigated the performance improvement of LSRA, we found that the reason for the poor performance of ViT is that it has small Head Diversity, as shown in Fig.\ref{fig:2}. The reason why LSRA outperforms than ViT is that the introducing IB improves the Head Diversity and thus enables Transformers to have more comprehensive information. However, as we can see, the Head Diversity of LSRA's deep layers is still small, so we believe that the introducing IB of LSRA is not efficient enough. We speculate that the network's performance will be further improved if IB is introduced more efficiently to make the Head Diversity at the deep layers greater. To further improve the efficiency of introducing IB, we investigate why LSRA brings more IB to the network than ViT and why LSRA cannot bring IB to the leaning back layers. Since the main structure of LSRA is the Invariant Embedded Parallel Structure (Invariant Structure), we first investigate it.

\subsection{Why Does It Work?}
In comparison with ViT, LSRA adds a convolution module in parallel at the MHA, as shown in Fig.\ref{fig:MainStruofViTandLSRA}. The Convs (i.e., IB Structure) and MHAs each process half of the data along the channel dimension. Since its structure is invariable, all the layers repeat the same operation.

Since the MHAs are the low-pass filters, they reduce the high-frequency data (HFD) share layer by layer \cite{park2022vision}. Because the HFD share drops with each layer, the deep MHAs have difficulty detecting the HFD,i.e., the short-distance information. As a result, the attention distances of ViT converge to a greater value as the layers deepen. The fact that the deep layers do not adequately utilize the HFD, we believe, is the reason of ViT's poor performance.

In comparison with ViT, LSRA introduces the convolutions, the hight-pass filters \cite{park2022vision}, to improve the HFD share. As a result, LSRA's HFD share drops more slowly, which makes it more use of the HFD, resulting in slower convergence of attention distance. Also, the improved HFD utilization leads to improved performance.

\subsection{Why Is It Suboptimal?}
You may now ask: why in a structure like Embedded Parallel, where MHAs and Convs each process half of the data along the channel dimension, does it still cause a drop in HFD share? We are confident that it is because the Receptive Field of Convs become larger as the layers increase \cite{krizhevsky2012imagenet}.

We know that HFD corresponds to local information, which requires a small Receptive Field. When the Receptive Field of Conv increases, its high-pass filtering ability decreases. For the network, it means inefficient introduction of IB. In contrast, MHA is a global attention receptor, and its Receptive Field is the whole feature map. The receptive field of MHA does not change as the layers increase, i.e., its low-pass filtering ability does not become weaker as the layers increase. Since the high-pass filtering ability of Conv gradually becomes weaker and the low-pass filtering ability of MHA remains the same, the overall HFD share decreases as a result.

\subsection{How to Do?}
To increase the HFD share of the deeper layers, an intuitive idea is to increase the share of data processed by the deeper Convs, called Increasing Structure. In the Increasing Structure, the deep MHAs share is decreasing, and the Convs share is increasing. The decrease of deep MHAs share reduces low-frequency data (LFD), and the increase of deep Convs share does not increase HFD (we will discuss this issue later). As a result, this structure increases the HFD share of the data by reducing LFD. However, our ultimate goal is to give MHAs more comprehensive information, so we cannot increase the HFD share by reducing LFD.

\begin{figure}[htb]
\centering
\includegraphics[width=0.9\textwidth]{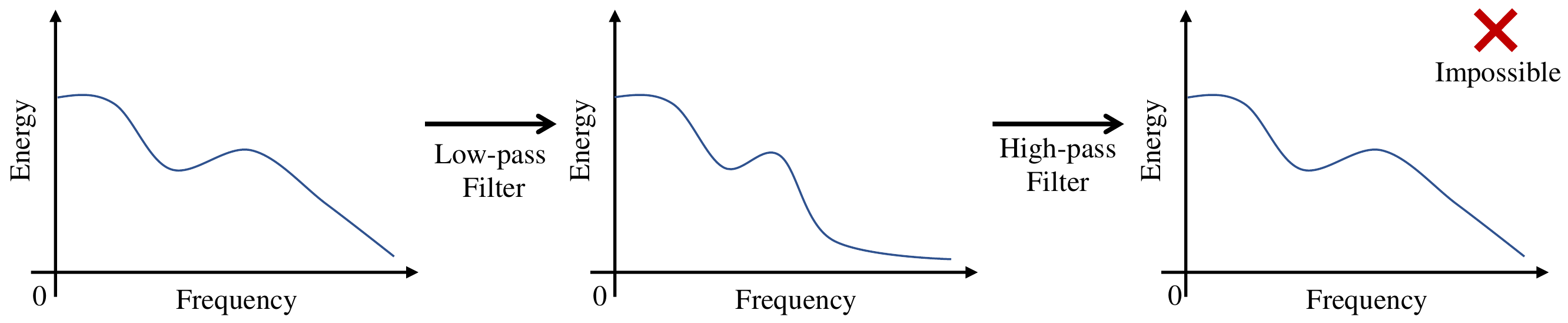}
\caption{Low/High-Pass Filter Processes Data.}
\label{fig:LowHighPassFilter}
\vspace{-0.25cm}
\end{figure}

Why the increase of deep Convs share does not increase HFD? Because the HFD has been filtered out by the shallow layers. We can not use a low-pass filter to filter out the high-frequency components of the data and then use a high-pass filter to extract its high-frequency components. This is impossible because the high-frequency components have already been removed, and you cannot get the components that the high-pass filter has removed, as shown in Fig.\ref{fig:LowHighPassFilter}.

We think a reasonable approach is to increase the proportion of Conv processed data in the shallow layers and keep as much HFD as possible for the deep layers. At the same time, decreasing the proportion of MHA processed data in the shallow layers to reduce the weakening ability of MHA on HFD. As a result, making more HFD can be passed to the deeper layers of the network. Of course, we cannot just increase the HFD share in the network because our goal is to give MHA more comprehensive information rather than just HFD. For this reason, we also increase the proportion of data processed by the MHA in the deeper layers.

\subsection{Network Architecture}
\begin{figure}[htb]
\centering
\includegraphics[width=\textwidth]{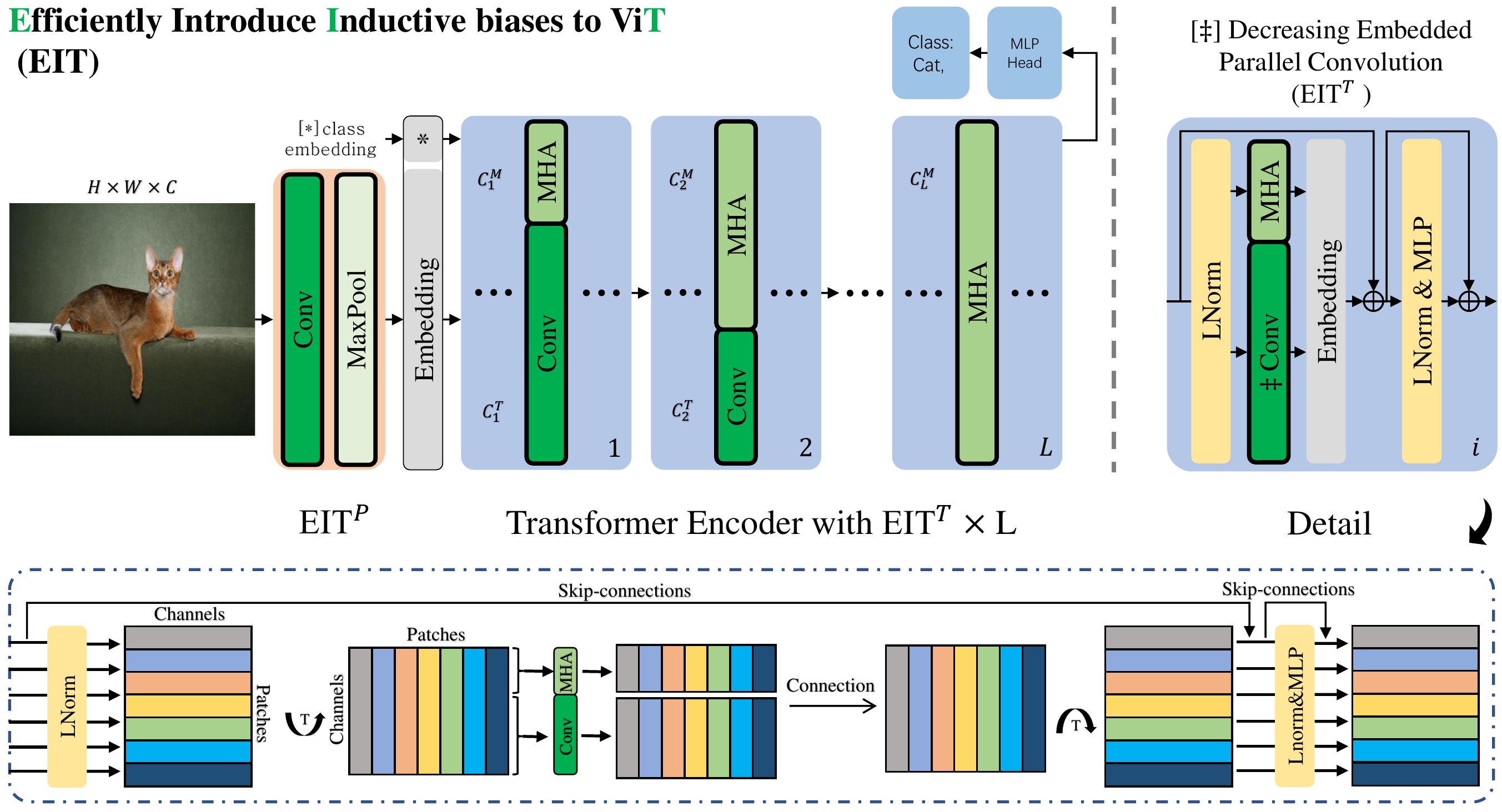}
\caption{Model Overview. The backbone of EIT is the same as ViT. EIT$^T$ and MHA each process all the patches embedding in \textbf{different channel dimensions}. $C_i^{T}$ and $C_i^{M}$ are the number of channels processed by EIT$^T$ and MHA, respectively.}
\label{fig:3}
\vspace{-0.6cm}
\end{figure}

\subsubsection{EIT}
Based on the above analysis, we design a Decreasing Structure to improve the efficiency of the IB introduction. The architecture of the proposed model is shown in Fig.\ref{fig:3}. We propose two simple structures called EIT$^P$ (EIT for PaPr) and EIT$^T$ (EIT for TrEn) to form a complete Decreasing Structure.

\subsubsection{EIT for PaPr}
EIT$^P$ uses a convolution layer with a stride smaller than the kernel size, i.e., there are overlapping patches, which can improve the locality, result in introducing IB efficiency of PaPr. However, it also means we will introduce redundant patches to the TrEns, introducing excess FLOPs. To solve this problem without seriously destroying the similarity of the adjacent patches, we filter out the redundant patches by a maximum pooling layer.

\subsubsection{EIT for TrEn}
The structure of introducing IB in LSRA \cite{Wu2020Lite} contains one activation layer, one convolutional layer, and one fully connected layer. EIT$^T$ contains only one layer of convolution. Because we believe that if we aim to use the property of the high-pass filter of the convolution, we only need to have one convolutional layer. Multiple convolutional layers increase the Receptive Field and weaken the EIT$^T$'s high-pass filter ability. Other types of layers (e.g., activation layers, fully connected layers) do not enhance EIT$^T$'s high-pass filtering ability, and we consider them dispensable.

\section{Experiments}

\subsection{Set up}
We use four popular small-scale datasets and one large-scale dataset to evaluate the performance of EIT: Cifar10/100 \cite{Krizhevsky2009Learning}, Fashion-Mnist \cite{Xiao2017Fashion}, Tiny ImageNet-200 \cite{Tavanaei2020Embedded}, and ImageNet-1K \cite{krizhevsky2012imagenet}. The model configuration is detailed in Table \ref{tab:2} and the training configuration is in Table \ref{tab:imagenetTrain} and Table \ref{tab:3}, respectively. We train our model on ImageNet-1K with strong data augmentation. Specifically, we adopt the same data augmentations (RandAugment \cite{cubuk2020randaugment}, MixUp \cite{zhang2017mixup}, CutMix \cite{yun2019cutmix} and Erasing \cite{zhong2020random}) as DeiT \cite{Touvron2021Training}.

\begin{table}[ht]
\small
\centering
\fontsize{5.8}{8}\selectfont
\caption{Details of EIT model variants. EIT16/4/3 indicates that the convolutional kernel size used in EIT$^P$ is 16, the stride of EIT$^P$ is 4, and the kernel size and stride of the maxpool in EIT$^P$ is 3, and C is the number of channels (i.e., embedding dimension). The kernel size and stride of EIT$^T$ are 3 and 1, respectively, by default. The "Params" is for a 1000-category classification task with an input image size of 224$\times$224. We utilize standard convolutions in EIT$^P$ and depth-wise separable convolution \cite{howard2017mobilenets} in EIT$^T$.}\label{tab:2}
\setlength{\tabcolsep}{0.3mm}{
\begin{tabular}{|c|c|c|c|c|c|c|c|}
\hline
Model                                                  & C   & EIT$^P$                                                     & EIT$^T$                                                   & Layers & \begin{tabular}[c]{@{}c@{}}MLP Size\end{tabular} & Heads & \begin{tabular}[r]{@{}r@{}}Params.\end{tabular}\\ \hline
\begin{tabular}[c]{@{}c@{}}EIT\textcolor{green}{16}/\textcolor{blue}{4}/\textcolor{cyan}{3}-Mini\end{tabular} & 250 & \multirow{4}{*}{\begin{tabular}[c]{@{}c@{}}$C$@Conv:(\textcolor{green}{16,16},\textcolor{blue}{4});\\Maxpool:(\textcolor{cyan}{3,3,3})\end{tabular}}       & \multirow{4}{*}{\begin{tabular}[c]{@{}c@{}}$C^{T}$@\\Conv:(3,3,1,groups=$C^{T}$)\end{tabular}} & 5      & \multirow{4}{*}{4$C$}                                & 10    & 3.5M  \\
\begin{tabular}[c]{@{}c@{}}EIT\textcolor{green}{16}/\textcolor{blue}{4}/\textcolor{cyan}{3}-Tiny\end{tabular} & 330 &   &                                                                                 & 8      &                                                    & 10    & 8.9M  \\
\begin{tabular}[c]{@{}c@{}}EIT\textcolor{green}{16}/\textcolor{blue}{4}/\textcolor{cyan}{3}-Base\end{tabular} & 400 &  &                                                                                 & 10      &                                                    & 16    & 16.0M\\
\begin{tabular}[c]{@{}c@{}}EIT\textcolor{green}{16}/\textcolor{blue}{4}/\textcolor{cyan}{3}-Large\end{tabular} & 464 &  &                                                                                 & 12      &                                                    & 16    & 25.3M\\
\hline
\end{tabular}}
\vspace{-0.4cm}
\end{table}

\begin{table}[ht]
\centering
\fontsize{6}{7}\selectfont
\setlength{\tabcolsep}{0.9mm}{
\begin{tabular}{|c|c|c|c|c|c|c|c|}
\hline
Optimizer& Aug. & Scheduler & LR & LR min & Drop & Att. Drop & Droppath \\ \hline
SGD      & H. flip & Cosine    & 1e-3     & 1e-5    & 0.2  & 0.15      & 0.2      \\
\hline
\end{tabular}
}
\caption{Details of model training on the small-scale datasets. All the models are trained with a total batch size of 25 for 300 epochs.}\label{tab:3}
\vspace{-0.4cm}
\end{table}

\begin{table}[ht]
\centering
\fontsize{6}{7}\selectfont
\setlength{\tabcolsep}{0.75mm}{
\begin{tabular}{|cccccccccccccccccc|}
\hline
\rotatebox{90}{Optimizer}  &\rotatebox{90}{LR} &\rotatebox{90}{LR decay}  &\rotatebox{90}{LR min}  &\rotatebox{90}{Weight decay} &\rotatebox{90}{Warmup epochs} & \rotatebox{90}{Label smoothing}  & \rotatebox{90}{Dropout}& \rotatebox{90}{Stoch. Depth }& \rotatebox{90}{Repeated Aug}& \rotatebox{90}{H. flip}& \rotatebox{90}{RRC}& \rotatebox{90}{Rand Augment} & \rotatebox{90}{Mixup alpha} & \rotatebox{90}{Cutmix alpha} & \rotatebox{90}{Erasing prob.} & \rotatebox{90}{Test crop ratio}& \rotatebox{90}{Loss}\\ \hline
 AdamW& 1e-3& cosine& 1e-5& 0.05& 5& 0.1& 0& 0& 0& $\checkmark$& $\checkmark$& 9/0.5& 0.8& 1.0& 0.25 &0.875&  CE\\        \hline
\end{tabular}}

\caption{Details of model training on the ImageNet-1K. We train our models on 4 $\times$ V100-32G with a total batch size of 1024, 1024, 512 and 512 for EIT-Mini, EIT-Tiny, EIT-Base and EIT-Large, respectively, for 300 epochs.}
 \label{tab:imagenetTrain}
\end{table}

\subsection{Comparision on the ImageNet-1K}

\begin{figure}[htb]
\centering
\includegraphics[width=1\textwidth]{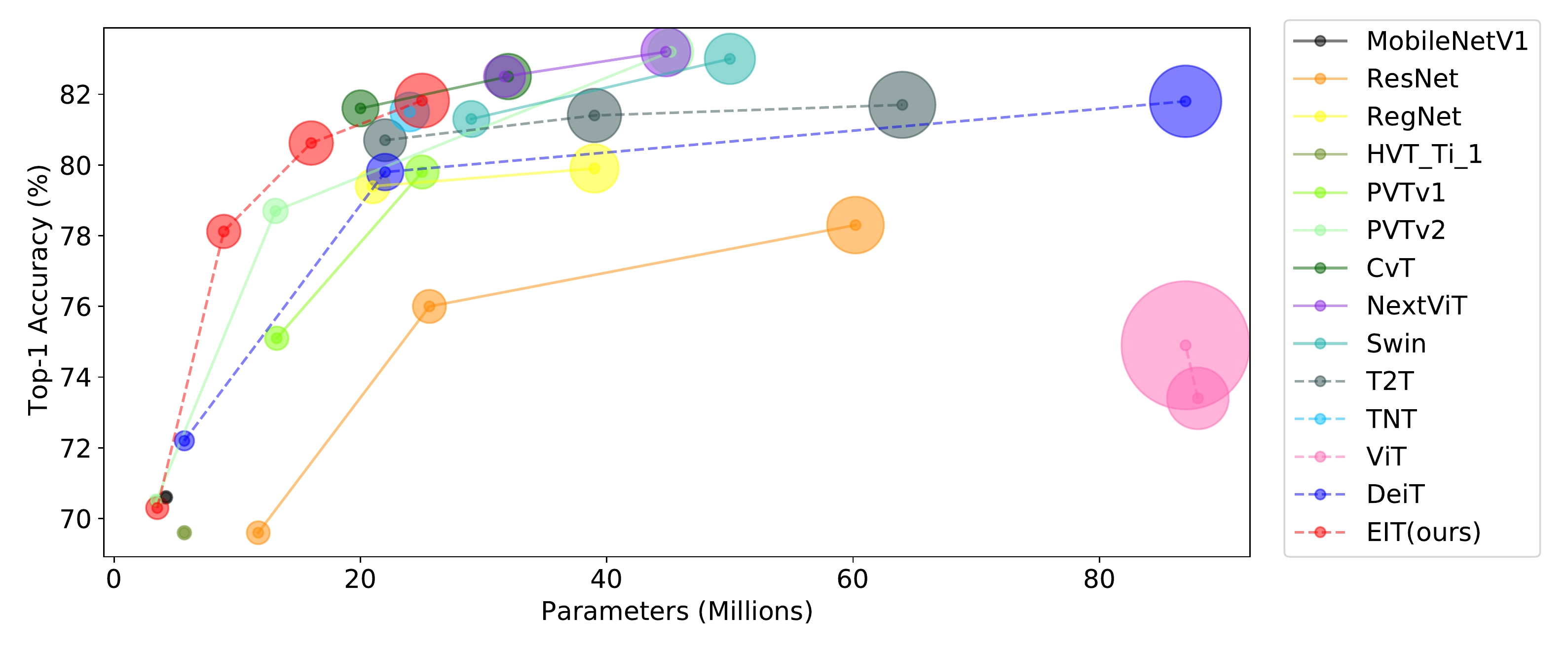}
\caption{Comparison on the ImageNet-1K. The size of the circle corresponds to FLOPs. The solid line and dashed line represent the model with pyramid structure and the model with pyramid-free structure, respectively. Existing networks with a pyramidal structure are generally better than networks with a pyramid-free structure. Our EIT beats the pyramid structure with a pyramid-free structure. Because of the efficient use of comprehensive-frequency data by EIT, the parameter utilization of EIT is the highest.}
\label{fig:Imagenet1kAcc}
\vspace{-0.5cm}
\end{figure}

Table \ref{tab:imagenet} and Fig.\ref{fig:Imagenet1kAcc} discusse EIT's performance on the ImageNet-1K with other state-of-the-art methods.

When compared to MobileNetV1 \cite{howard2017mobilenets}, EIT16/4/3-Mini obtains 70\% Top-1 accuracy with 83\% parameters in models with pyramid structure. In comparison to ResNet152 \cite{He2016Deep}, EIT16/4/3-Tiny achieves 78\% Top-1 accuracy with 15\% parameters and 35\% FLOPs. EIT16/4/3-Base obtains 80.6\% Top-1 accuracy, which is 0.7\% higher than RegNetY-8G \cite{radosavovic2020designing}, with 41\% parameters and 82\% FLOPs. EIT16/4/3-Mini obtains 0.4\% higher Top-1 accuracy with 61\% parameters than HVT-Ti-1 \cite{pan2021scalable}. EIT16/4/3-Tiny provides a similar Top-1 accuracy with 68\% parameters as PVTv2-B1 \cite{wang2022pvt}. EIT16/4/3-Base obtains the similar Top-1 accuracy, with 80\% and 55\% parameters, respectively, as CvT-13 \cite{Wu2021CvT} and Swin-T \cite{liu2021swin}. EIT16/4/3-Large provides a similar Top-1 accuracy with 78\% and 80\% parameters, respectively, as CvT-21 \cite{Wu2021CvT} and Next-ViT-S \cite{li2022nextvit}.

In models with a pyramid-free structure, EIT16/4/3-Base obtains a similar Top-1 accuracy as T2T-14 \cite{Yuan2021Tokens} while reducing 27\% parameters. When compared to the TNT-S \cite{Han2021Transformer}, the EIT16/4/3-Base achieves the similar Top-1 accuracy with 67\% parameters. EIT16/4/3-Tiny achieves 3.2\% higher Top-1 accuracy with 7\% FLOPs and 10\% parameters than ViTB/16 \cite{Dosovitskiy2021An}. Furthermore, compared to DeiT-B \cite{Touvron2021Training}, EIT16/4/3-Large achieves the same Top-1 accuracy with 29\% parameters and 59\% FLOPs.

EIT achieves promising and competitive results than the similar representative state-of-the-art methods currently available. In particular, EIT achieves state-of-the-art performance over the other same scale models which have a pyramid-free structure. The models with pyramid structure generally contain fewer FLOPs and parameters than the models with pyramid-free structure. In terms of FLOPs, EIT closes the 'gap' between pyramid-free and pyramid structures and delivers better results in terms of parameters. This signifies that the pyramid structure is not required for image processing. The pyramid-free structures also keep the model brevity, facilitating the adjustment of model parameters.

\begin{table}[htbp]
\centering
\fontsize{5.5}{8}\selectfont
\setlength{\tabcolsep}{2.5mm}{
\caption{Comparison on the ImageNet-1K. The image size is $224\times224$.}\label{tab:imagenet}
\begin{tabular}{|c|c|c|c|c|c|}
\hline
Model       & Source                       & \begin{tabular}[c]{@{}c@{}}Pyramid\end{tabular}& \multicolumn{1}{c}{FLOPs} & Param                     & \begin{tabular}[c]{@{}c@{}}Top-1\end{tabular} \\ \hline

MobileNetV1 \cite{howard2017mobilenets}   & arXiv2017 &$\checkmark$                   & 0.58G & 4.2M                      & 0.706                                                    \\
\hline
ResNet18 \cite{He2016Deep}   & CVPR2016&      $\checkmark$                   & 1.82G                     & 11.7M & 0.696                                                                                                      \\
ResNet50 \cite{He2016Deep}   &  CVPR2016&       $\checkmark$                  & 3.77G                     & 25.6M & 0.760 \\
ResNet152 \cite{He2016Deep}   &CVPR2016&         $\checkmark$                  & 11G                     & 60.2M & 0.783                                                    \\ \hline
RegNetY-4G \cite{radosavovic2020designing}   &CVPR2020&         $\checkmark$                  & 4G                     & 21M & 0.794                                                    \\
RegNetY-8G \cite{radosavovic2020designing}   &CVPR2020&         $\checkmark$                  & 8G                     & 39M & 0.799                                                    \\ \hline
HVT-Ti-1 \cite{pan2021scalable}     & ICCV2021&  $\checkmark$                   & 0.64G & 5.7M                      & 0.696                                                    \\ \hline
PVTv1-Tiny \cite{wang2021pyramid}    & ICCV2021&  $\checkmark$                  & 1.90G  & 13.2M & 0.751                                                    \\
PVTv1-Small \cite{wang2021pyramid}    & ICCV2021&  $\checkmark$                  & 3.80G  & 25M & 0.798                                                    \\
PVTv2-B0 \cite{wang2022pvt}    & CVM2022&  $\checkmark$                  & 0.6G  & 3.4M & 0.705                                                    \\
PVTv2-B1 \cite{wang2022pvt}    & CVM2022&  $\checkmark$                  & 2.1G  & 13.1M & 0.787                                                    \\
PVTv2-B3 \cite{wang2022pvt}    & CVM2022&  $\checkmark$                  & 6.9G  & 45.2M & 0.832                                                    \\   \hline
CvT-13 \cite{Wu2021CvT}  & ICCV2021&  $\checkmark$                  & 4.5G  & 20M                      & 0.816                                                    \\
CvT-21 \cite{Wu2021CvT}  & ICCV2021&  $\checkmark$                  & 7.1G  & 32M                      & 0.825                                                    \\ \hline
Swin-T \cite{liu2021swin}  &ICCV2021&   $\checkmark$                  & 4.5G  & 29M                      & 0.813                                                    \\
Swin-S \cite{liu2021swin}  & ICCV2021&  $\checkmark$                  & 8.7G  & 50M                      & 0.830                                                    \\ \hline
Next-ViT-S  \cite{li2022nextvit}  & ECCV2022&      $\checkmark$                & 5.8G  & 31.7M                      & 0.825 \\
Next-ViT-B  \cite{li2022nextvit}  & ECCV2022&      $\checkmark$                & 8.3G  & 44.8M                      & 0.832                                                    \\ \hline
T2T-14 \cite{Yuan2021Tokens}  &ICCV2021&   $\times$                  & 6.1G  & 22M                      & 0.807                                                    \\
T2T-19 \cite{Yuan2021Tokens}  &ICCV2021&   $\times$                  & 9.8G  & 39M                      & 0.814                                                    \\
T2T-24 \cite{Yuan2021Tokens}  &ICCV2021&   $\times$                  & 15G  & 64M                      & 0.817                                                    \\\hline
TNT-S \cite{Han2021Transformer}  &NIPS2021 &  $\times$                  & 5.2G  & 24M                      & 0.815                                                    \\\hline
ViTB/32 \cite{Dosovitskiy2021An}    &ICLR2021&  $\times$                     & 13G   & 88M                       & 0.734                                                    \\
ViTB/16 \cite{Dosovitskiy2021An}    &ICLR2021&  $\times$                      & 56G   & 87M                       & 0.749                                                    \\ \hline
DeiT-T \cite{Touvron2021Training}  &ICML2021&   $\times$                  & 1.3G  & 5.7M                      & 0.722                                                    \\
DeiT-S \cite{Touvron2021Training}  &ICML2021&   $\times$                  & 4.6G & 22M                      & 0.798   \\
DeiT-B \cite{Touvron2021Training}  &ICML2021&   $\times$                  & 17.5G  & 87M                      & 0.818                                                  \\

\hline

EIT16/4/3-Mini &-&$\times$& 1.73G                     & \textbf{3.5M}                      & 0.700                                                    \\
EIT16/4/3-Tiny &-&$\times$& 3.84G                     &\textbf{8.9M} & 0.781                                                    \\
EIT16/4/3-Base &-&$\times$& 6.52G                          & \textbf{16.0M}      & 0.806                                     \\
EIT16/4/3-Large &-&$\times$& 10.0G                          & \textbf{25.3M}     & 0.818                                     \\
\hline
\end{tabular}
}
\vspace{-0.5cm}
\end{table}

\subsection{Comparision on the small-scale datasets}
In this section we discuss the performance of CvT \cite{Wu2021CvT}, LSRA \cite{Wu2020Lite}, ViT$_C$ \cite{Xiao2021Early} and EIT in terms of IB introduction for the same architecture based on small-scale datasets. In the following experiments, we use Model Idx to denote the model.

We discuss the performance of EIT based on EIT3/1/4-Mini. We compare how EIT introduces IB in both phases (PaPr and TrEn) of ViT with that of CvT \cite{Wu2021CvT}, LSRA \cite{Wu2020Lite} and ViT$_C$ \cite{Xiao2021Early}, with the same network architecture. The results are shown in Table \ref{tab:4}, showing that both structures of EIT exhibit more efficient introducing IB. On the four datasets, compared with ViT, EIT has the average improvement of 12.6\% with fewer parameters and FLOPs. Compared with CvT, LSRA and ViT$_C$, the average improvement of EIT are 6.4\%, 7.3\% and 10.7\%, respectively.

\begin{table}[htbp]
\centering
\caption{Comparison with the ViTs. The structural parameters (C, Layers, MLP size and Heads) of ViTs are the same. All the convolution of ViTs in PaPr is standard convolution. * We did not use the pyramid structure of the models to ensure the fairness of the comparison of ViTs. The convolutional kernel size of CvT$^P$ is 4, and the stride is 2. The value of $n$ in ViT$_C$ is 2, the convolutional kernel size is 4, and the stride is 2.}
\fontsize{5}{8}\selectfont
\setlength{\tabcolsep}{0.5mm}{
\begin{tabular}{|c|c|c|c|c|c|c|c|c|c|c|}
\hline
\multirow{2}{*}{\begin{tabular}[c]{@{}c@{}}Model\\Idx\end{tabular}} & \multirow{2}{*}{\begin{tabular}[c]{@{}c@{}}Method\\(ViTs)\end{tabular}}                                                          & \multirow{2}{*}{\begin{tabular}[c]{@{}c@{}}Intro. IB to \\PaPr Phase\end{tabular}} & \multirow{2}{*}{\begin{tabular}[c]{@{}c@{}}Intro. IB to \\TrEn Phase\end{tabular}}      & FLOPs & \begin{tabular}[c]{@{}c@{}}Pram.\end{tabular} &Cifar10 & Cifar100 & \begin{tabular}[c]{@{}c@{}}Fashion\\ Mnist\end{tabular} & \begin{tabular}[c]{@{}c@{}}Tiny \\ImageNet\end{tabular} & \multirow{2}{*}{Avg.}  \\ \cline{5-10}
                      &                                                                                  &                                                                                              &  &                                                                                                 \multicolumn{6}{c}{\begin{tabular}[c]{@{}c@{}}Image Size:32,32 ;Patch Size:4\end{tabular}}                         &   \\ \hline
1.1                   & ViT \cite{Dosovitskiy2021An}                                                                              & None                                                                                         & None                                                                                              & 0.515G      & 3.798M &0.682&0.413&0.888&0.246&0.557(+0.0\%)             \\ \cline{1-11}
1.2                   & LSRA \cite{Wu2020Lite}                                                                            & None                                                                                         & LTB                  & 0.390G      & 2.924M &0.778&0.477&0.910&0.276&0.610(+5.3\%)                                         \\ \cline{1-11}
1.3                   & \multirow{3}{*}{CvT* \cite{Wu2021CvT}}                                                            & \begin{tabular}[c]{@{}c@{}}CvT$^P$\end{tabular}                    & None                                                                                              & 2.296G      & 3.846M &0.738&0.452&0.899&0.280&0.592(+3.5\%)                                                  \\
1.4                   &                                                                                  & None                                                                                         & \begin{tabular}[c]{@{}c@{}}CvT$^T$\end{tabular}    & 0.763G       & 5.706M & 0.712        & 0.414         & 0.909                                                        & 0.233     & 0.567(+1.0\%)                                                  \\
1.5                   &                                                                                  & \begin{tabular}[c]{@{}c@{}}CvT$^P$\end{tabular}                  & \begin{tabular}[c]{@{}c@{}}CvT$^T$\end{tabular}     & 3.277G      & 5.755M &0.777        & 0.497         & 0.914                                                        & 0.288              &0.619(+6.2\%)                                                  \\ \cline{1-11}
1.6                   &ViT$_C$*\cite{Xiao2021Early}                                                                                 & \begin{tabular}[c]{@{}c@{}}CoSt\end{tabular}                  & \begin{tabular}[c]{@{}c@{}}None\end{tabular}     & 0.553G & 4.108M &0.700        &0.431          &0.902                                                         &0.269    &0.576(+1.9\%)                                               \\ \cline{1-11}

1.7                   & \multirow{3}{*}{\begin{tabular}[c]{@{}c@{}}EIT3/1/4\\-Mini\end{tabular}} & \begin{tabular}[c]{@{}c@{}}EIT$^P$\end{tabular}      & None                                                                                              & 0.527G      & 3.793M  & 0.746        & 0.479         & 0.911                                                        & 0.283   &0.605(+4.8\%)                                                   \\
1.8                   &                                                                                  & None                                                                                         & \begin{tabular}[c]{@{}c@{}}EIT$^T$\end{tabular} & 0.501G       & 3.771M & 0.818        & 0.523         & 0.922                                                        & 0.313               &0.644(+8.7\%)                                                 \\
1.9                   &                                                                                  & \begin{tabular}[c]{@{}c@{}}EIT$^P$\end{tabular}      & \begin{tabular}[c]{@{}c@{}}EIT$^T$\end{tabular} & 0.428G      & 3.095M& \textbf{0.855}        & \textbf{0.605}          & \textbf{0.926}                                                        & \textbf{0.346}&\textbf{0.683(+12.6\%)}
\\
\hline
\end{tabular}
}
\label{tab:4}
\vspace{-0.5cm}
\end{table}

\begin{figure*}[htbp]
\centering
\begin{minipage}[b]{.3\linewidth}
\centering
\subfigure{\includegraphics[width=1\textwidth]{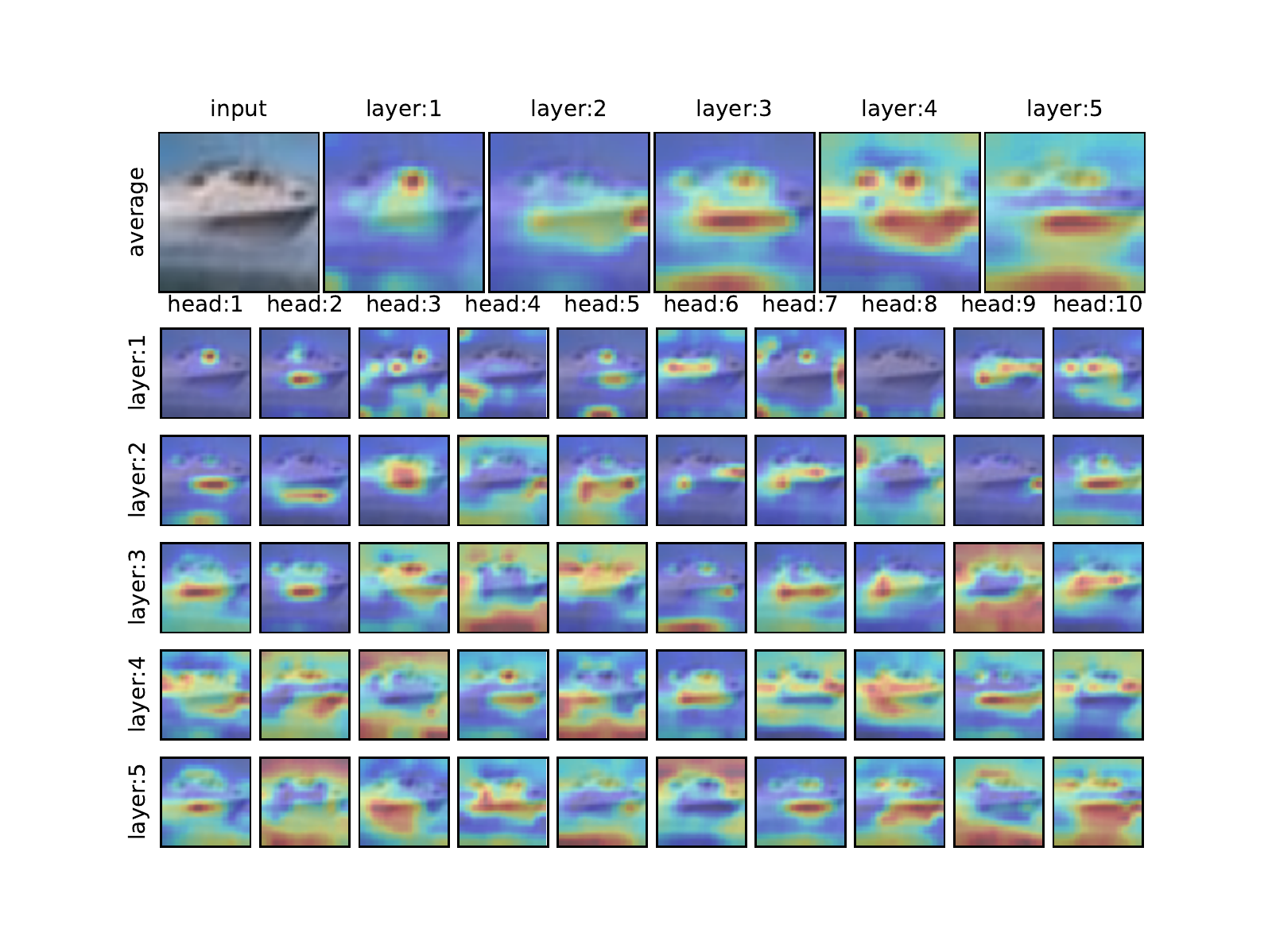}}
\vspace{-0.4cm}
\tiny{\center{~~~~~ViT \cite{Dosovitskiy2021An}.}}
\end{minipage}
\begin{minipage}[b]{.3\linewidth}
\centering
\subfigure{\includegraphics[width=1\textwidth]{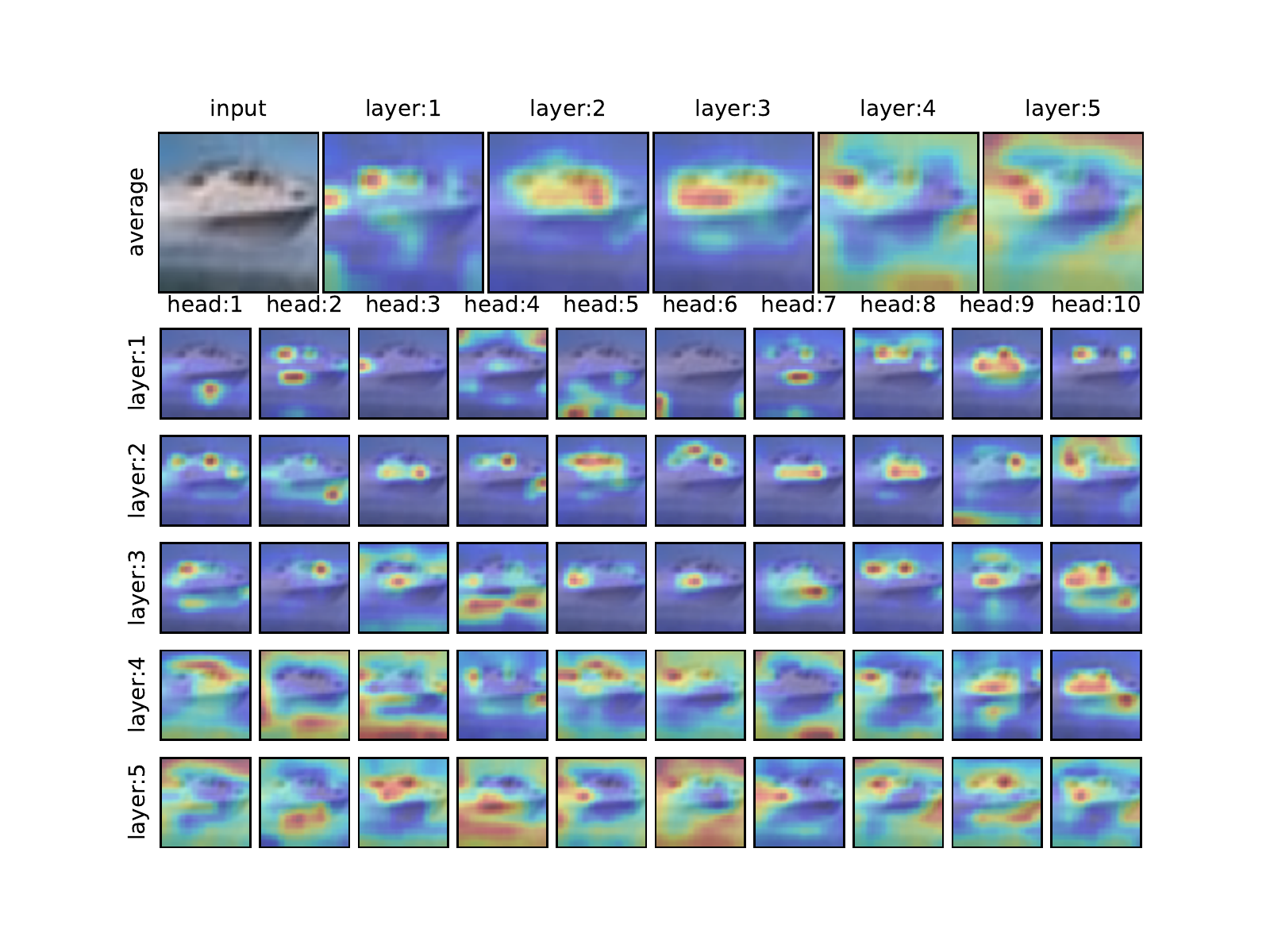}}
\vspace{-0.4cm}
\tiny{\center{~~~~~~LSRA \cite{Wu2020Lite}.}}
\end{minipage}
\begin{minipage}[b]{.3\linewidth}
\centering
\subfigure{\includegraphics[width=1\textwidth]{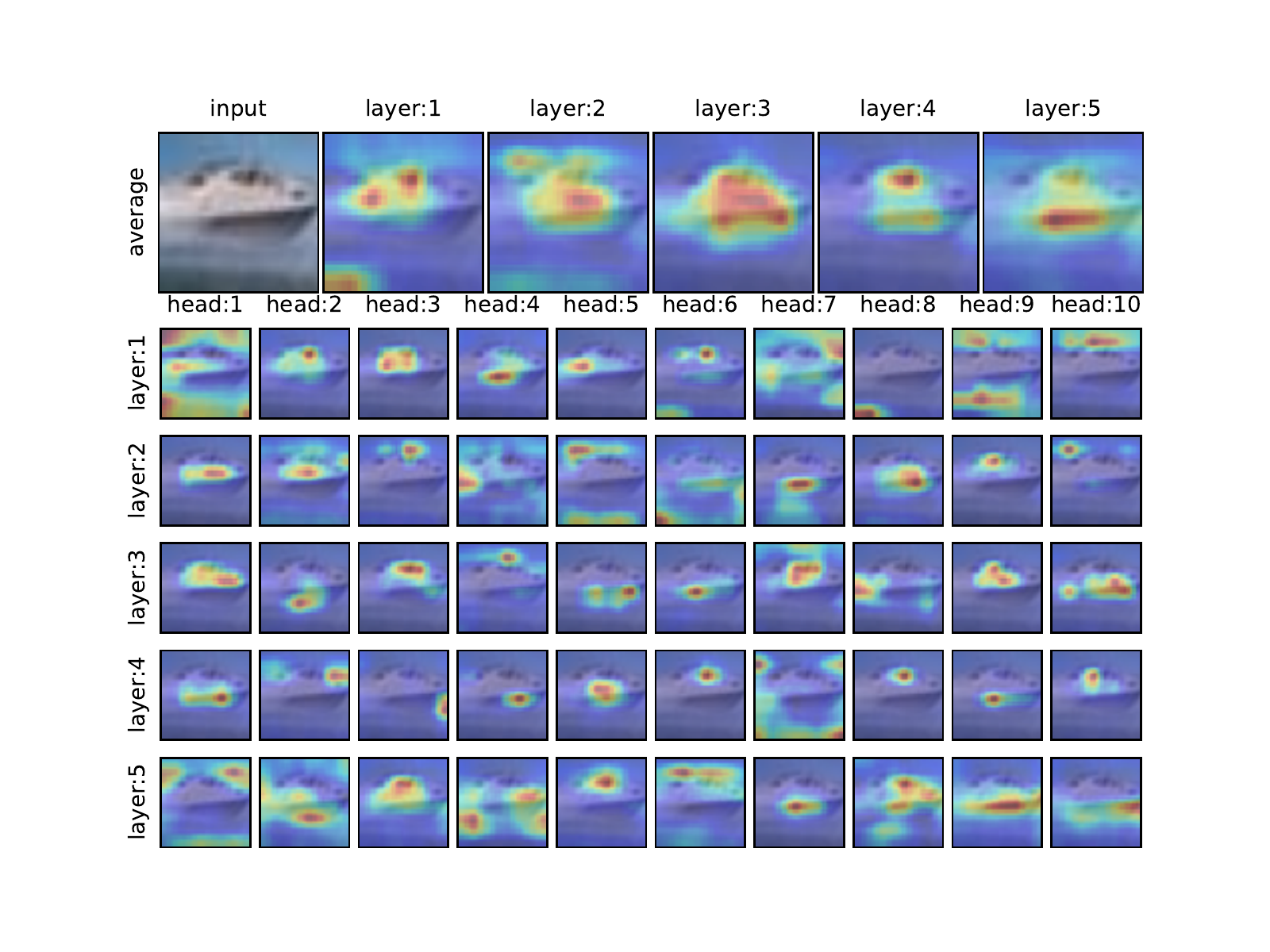}}
\vspace{-0.4cm}
\tiny{\center{~~~~~~EIT(ours).}}
\end{minipage}
\vspace{-0.3cm}
\caption{Attention Maps of ViTs. ViT's MHA attention range is expanding layer by layer. LSRA's attention range expands more slowly than ViT due to its improvement of HFD. Since the EIT guarantees the deep MHA's HFD, the attention can still focus on short-distance. This means the deep MHAs are not actively paying attention to the low frequencies (long-distance). The real reason is that they receive too little HFD compared with LFD. That is why the attention potential of ViT is not being fully exploited, resulting in poor performance. }
\label{fig:attentionMap}
\end{figure*}

\subsection{Visualization}

\begin{figure}[htbp]
\centering
\begin{minipage}[b]{.32\linewidth}
\centering
\vspace{-0.1cm}
\subfigure{\includegraphics[width=1\textwidth]{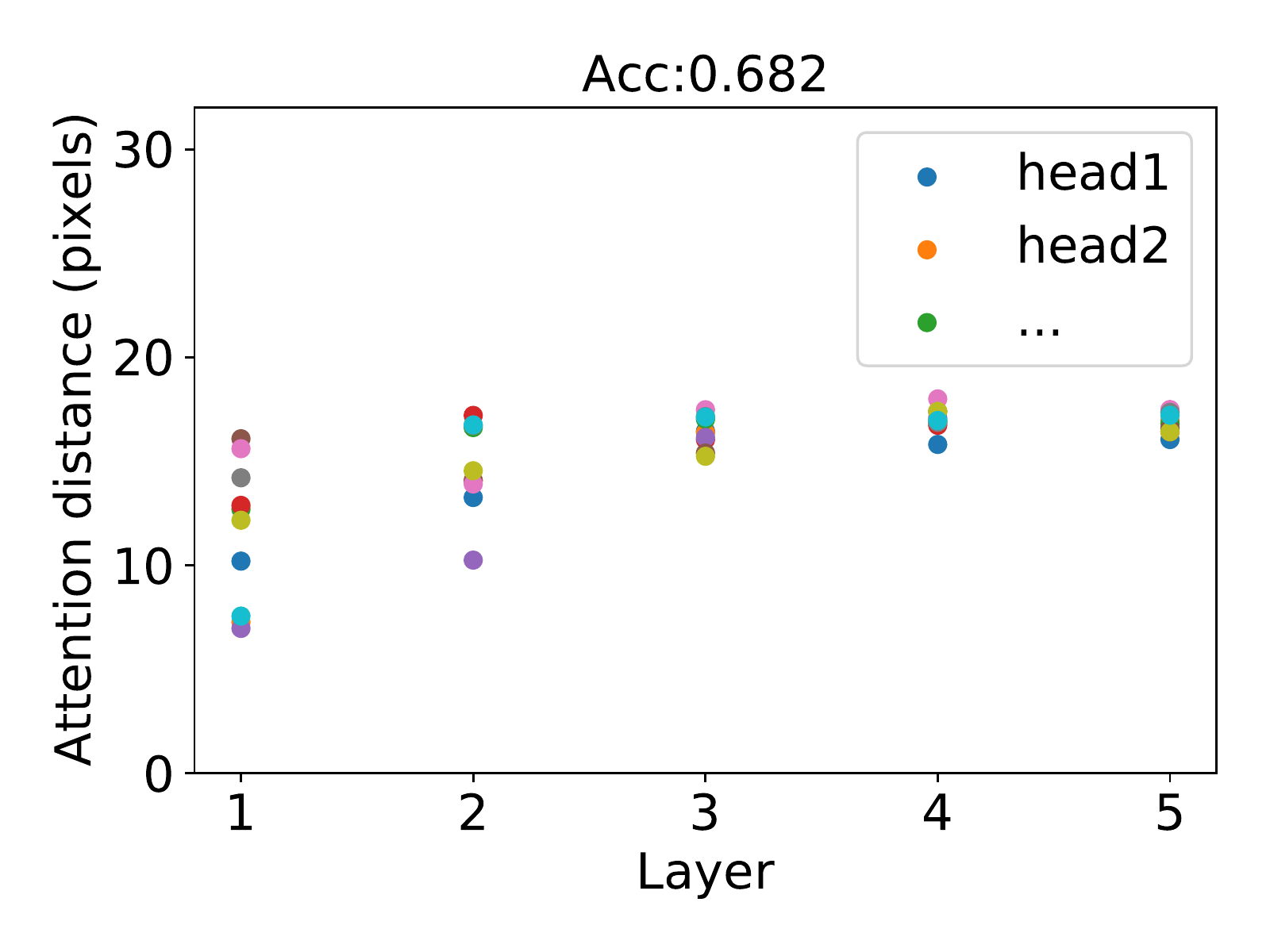}}
\vspace{-0.4cm}
\tiny{\center{~~~~~ViT \cite{Dosovitskiy2021An}}}
\vspace{-0.4cm}
\end{minipage}
\begin{minipage}[b]{.32\linewidth}
\vspace{-0.1cm}
\centering
\subfigure{\includegraphics[width=1\textwidth]{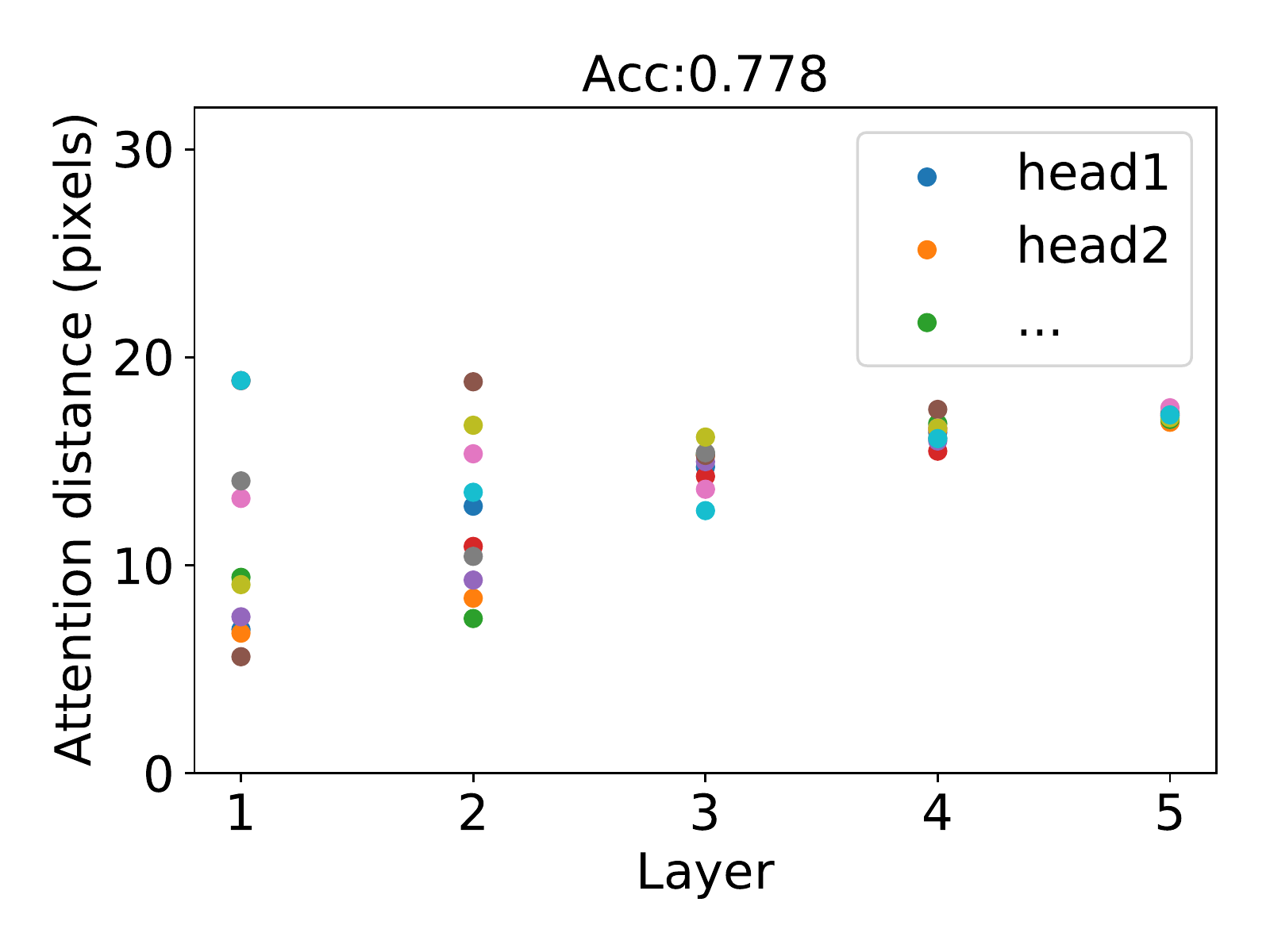}}
\vspace{-0.4cm}
\tiny{\center{~~~~~~LSRA \cite{Wu2020Lite}}}
\vspace{-0.4cm}
\end{minipage}
\begin{minipage}[b]{.32\linewidth}
\vspace{-0.1cm}
\centering
\subfigure{\includegraphics[width=1\textwidth]{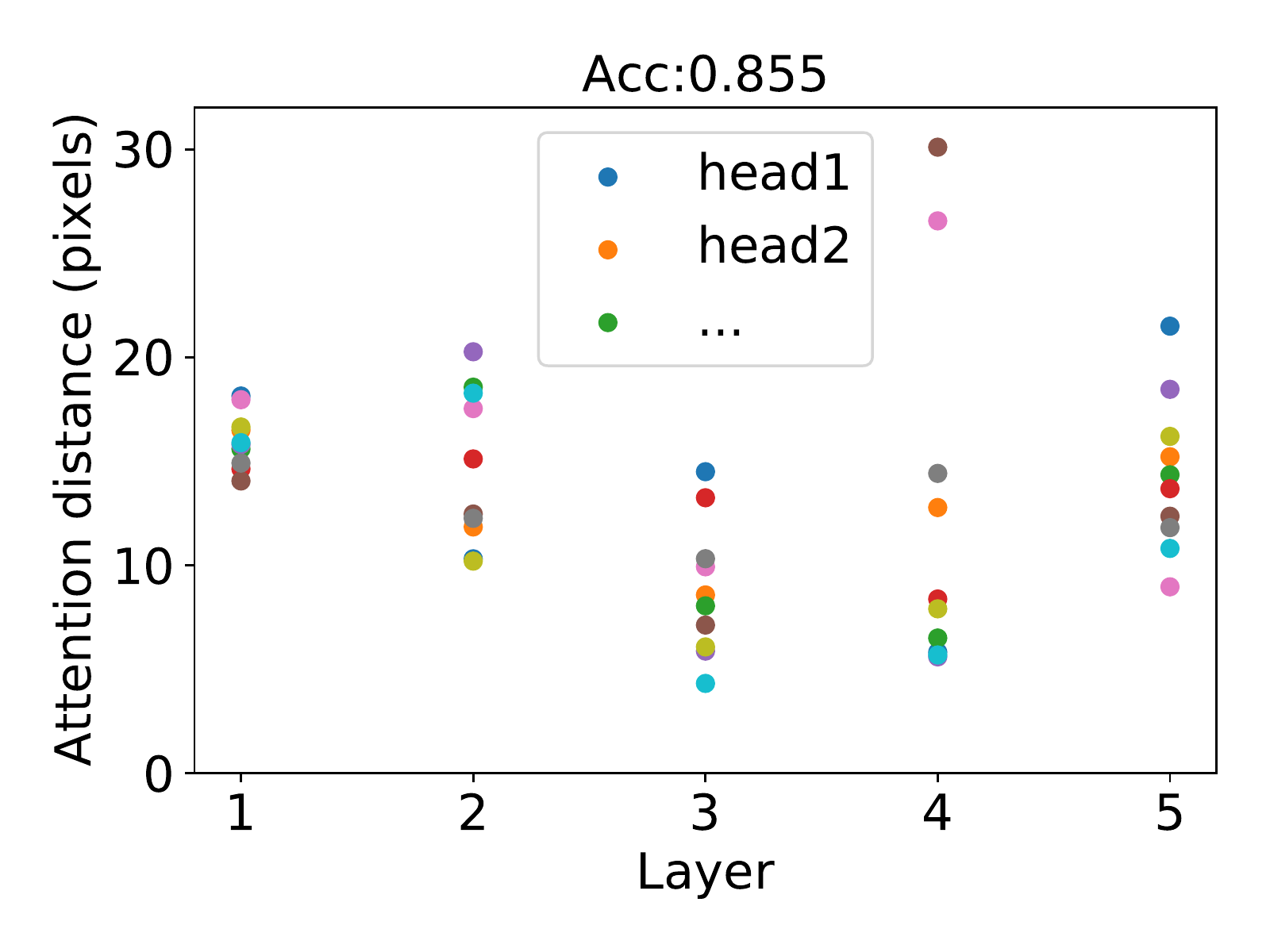}}
\vspace{-0.4cm}
\tiny{\center{~~~~~~EIT(ours)}}
\vspace{-0.4cm}
\end{minipage}

\caption{Attention Distance of ViTs. The top two layers of EIT have less Head Diversity than LSRA. However, the deep layers (layers 3, 4, and 5) have larger Head Diversity. Overall, EIT has a larger Head Diversity. It is apparent that increasing the network's Head Diversity improves performance. Furthermore, effective introducing IB can immediately boost deep layer Head Diversity. Compared with the other ViT and LSRA, EIT introduces IB more efficiently, resulting in better performance.}
\label{fig:4}
\vspace{-0.15cm}
\end{figure}

To verify if the performance improvement of EIT is due to the improvement of Header Diversity of deep MHAs, we computed the attention distances of ViT, LSRA, and EIT, with the Model Idx 1.1, 1.2, and 1.9 respectively, as shown in Fig.\ref{fig:4}. The attention distances are obtained by the same operation mentioned in Fig.\ref{fig:2}.

\begin{figure}[htb]
\centering
\begin{minipage}[b]{.32\linewidth}
\centering
\vspace{-0.1cm}
\subfigure{\includegraphics[width=1\textwidth]{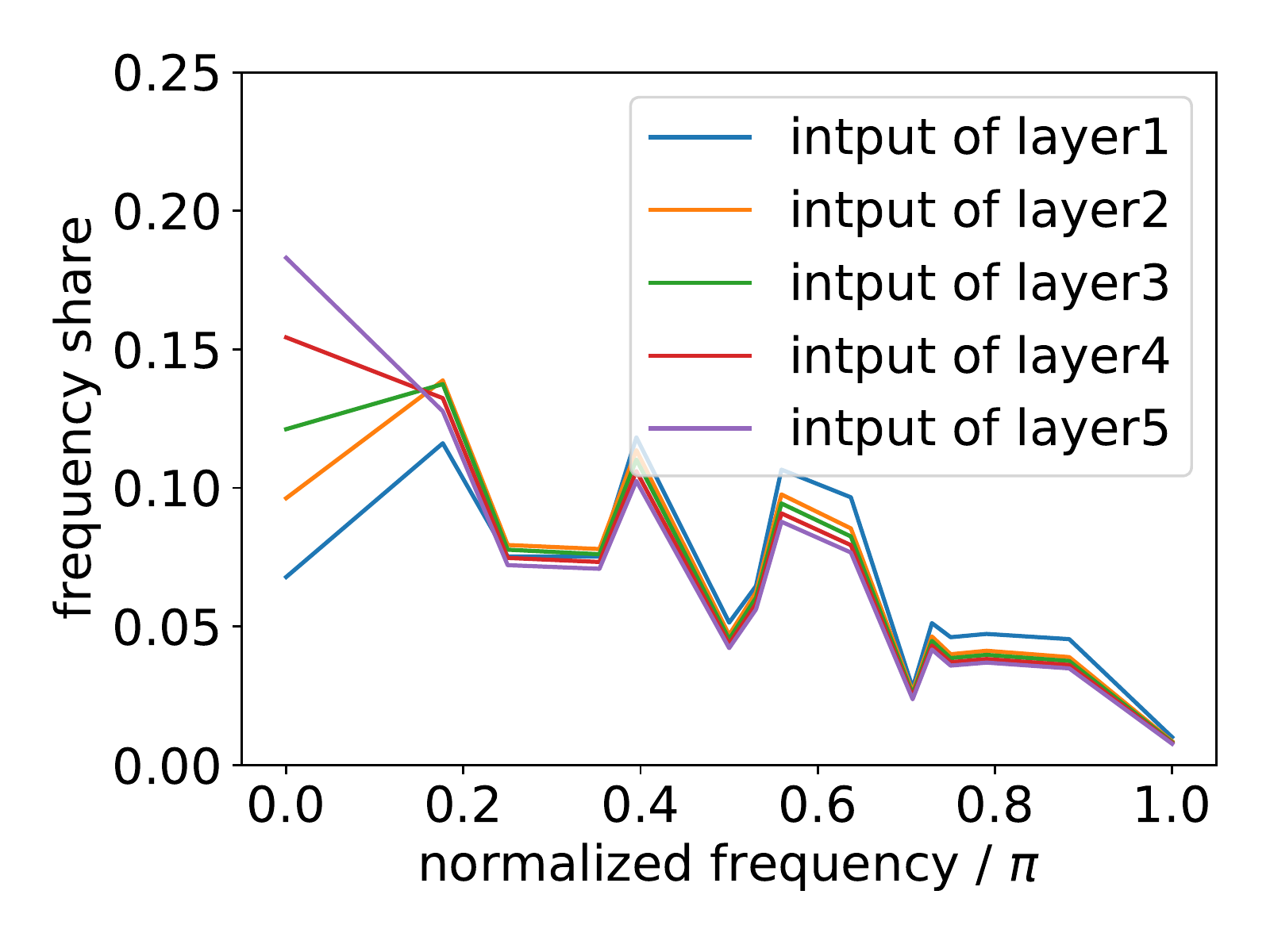}}
\vspace{-0.3cm}
\tiny{\center{~~~~~ViT \cite{Dosovitskiy2021An}}}
\vspace{-0.4cm}
\end{minipage}
\begin{minipage}[b]{.32\linewidth}
\vspace{-0.1cm}
\centering
\subfigure{\includegraphics[width=1\textwidth]{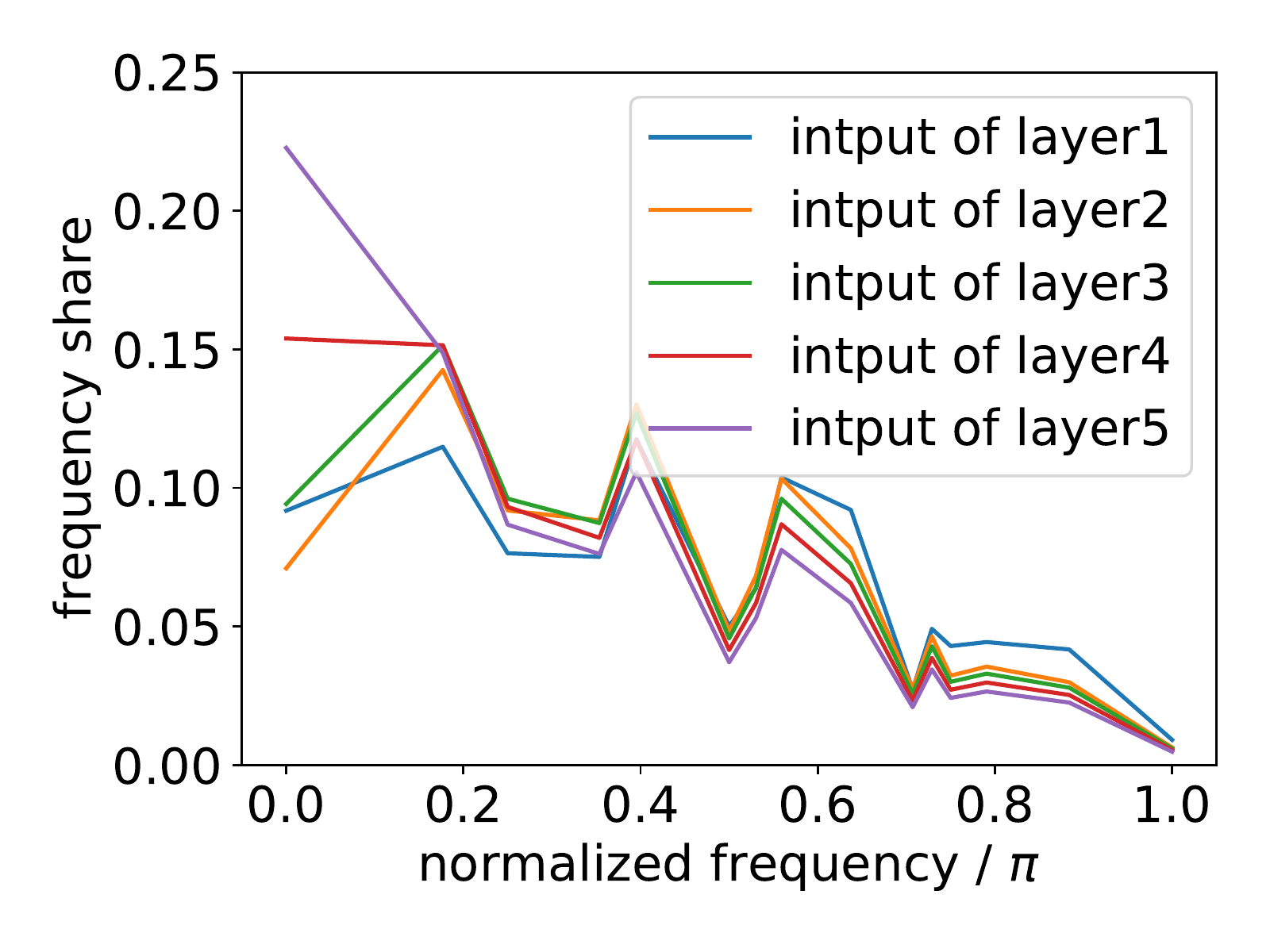}}
\vspace{-0.3cm}
\tiny{\center{~~~~~~LSRA \cite{Wu2020Lite}}}
\vspace{-0.4cm}
\end{minipage}
\begin{minipage}[b]{.32\linewidth}
\vspace{-0.1cm}
\centering
\subfigure{\includegraphics[width=1\textwidth]{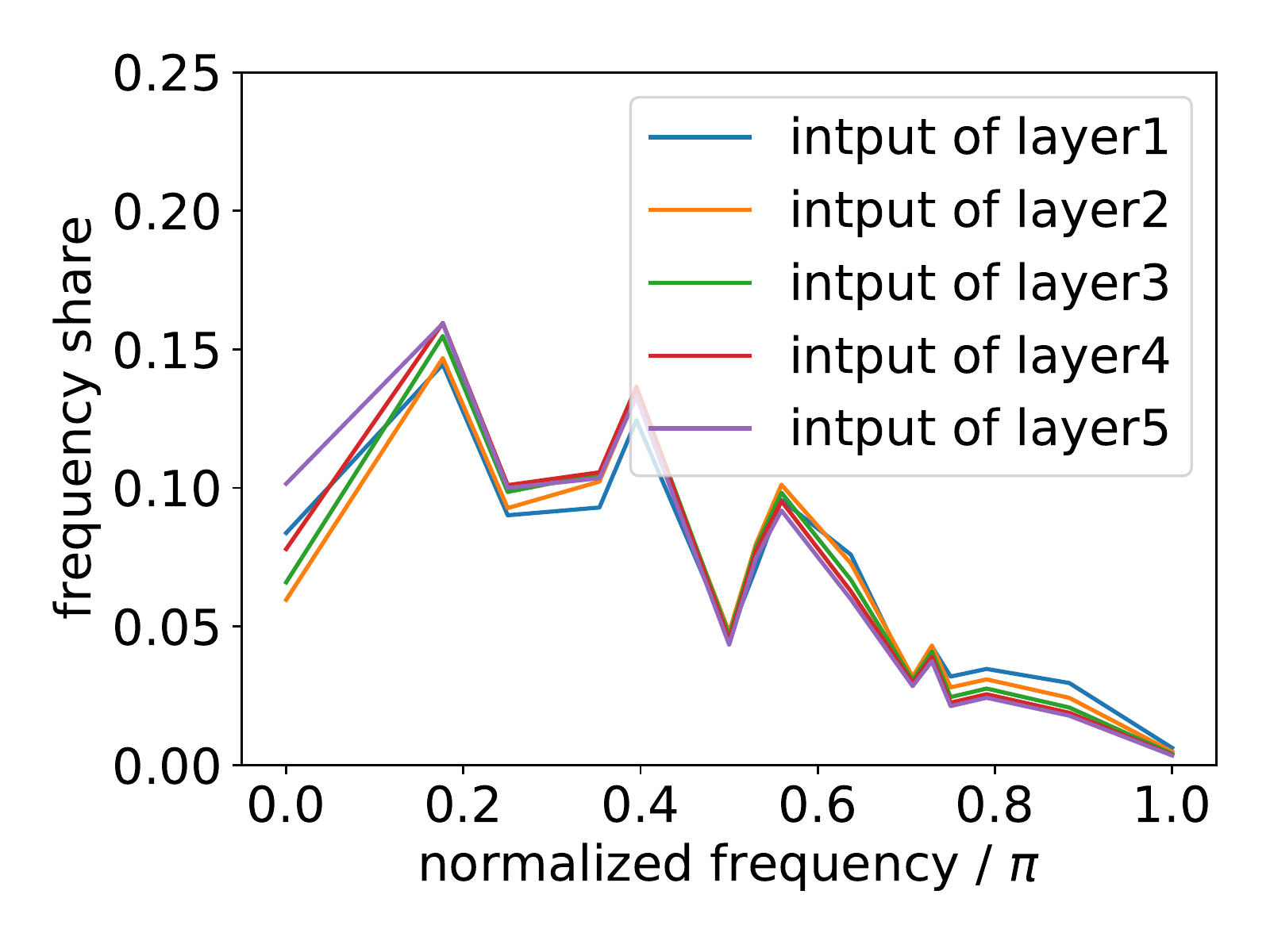}}
\vspace{-0.3cm}
\tiny{\center{~~~~~~EIT(ours)}}
\vspace{-0.4cm}
\end{minipage}

\caption{Frequency Distribution of ViT, LSRA, and EIT. ViT's HFD $(>0\pi)$ share is decreasing layer by layer, which is unfair to the deep MHAs. Compared with LFD, the deep MHAs can hardly notice HFD, which introduces to the limited attention mechanism of MHA, resulting in poor performance. Compared with ViT, the HFD share of LSRA tends to increase in the first three layers, so the MHAs can receive more HFD and shows a larger Head Diversity in Fig.\ref{fig:4}. However, the HFD share in the last two layers of LSRA is still decreasing, resulting in a small Head Diversity in Fig.\ref{fig:4}. Compared with ViT and LSRA, EIT has less reduction in HFD share per layer, ensuring an overall larger Head Diversity. It is worth noting that the Head Diversity of the EIT's first layer is smaller than ViT and LSRA in Fig.\ref{fig:4}. We believe this is the real self-attention (the deep layers' large Head Diversity is more important than the shallow layer) shown when each layer's attention mechanism is improved as much as possible.}
\label{fig:fre}

\end{figure}

We speculate in Section 3.2 that the poor performance of small Head Diversity is due to the smaller Header Diversity representing less HFD share in received data. To verify this speculation, we calculate the frequency share distribution of each layer's input by summing each channel's two-dimensional Fourier Transform for 2000 images of Cifar10 \cite{Krizhevsky2009Learning}. For the $i$-layer's input $x_{i} \in \mathbb{R}^{(1+H_{i}W_{i})\times{C_{i}}}$, the $j$-frequency share $\gamma_{ij}$ is defined as follows.
\begin{small}
\begin{eqnarray}
\gamma_{ij}=\frac{\sum_c|F_{j}(Reshape2D(x_i[1:,c]))|}{\sum_f\sum_c|F_f(Reshape2D(x_i[1:,c]))|}
\end{eqnarray}
\end{small}where $F(\cdot)$ is the two dimensional Fourier Transform, and $F_j(\cdot)$ represents the component with frequency $j$. $c$ and $f$ are the channel index and frequency index, respectively. The results are as shown in Fig.\ref{fig:fre}.

To visualize the effect of HFD on the MHA of each layer, we calculate the attention maps by randomly selecting one image from Cifar10, as shown in Fig.\ref{fig:attentionMap}. The attention maps of each layer are obtained by averaging that of heads. The attention maps of each head are the average of all channels.

\subsection{Ablation Study}
On the Cifar10/100 dataset, we designed four ablation experiments based on EIT3/1/4-Mini to verify that: 1) Decreasing Structure (EIT$^T$) works better than Parallel Structure (Fig.\ref{fig:2}-(c)); 2) the presence or absence of convolutional layers in ELT$^T$ is essential; 3) compared with Increasing and Invariant Structure, the Decreasing Structure is optimal; 4) EIT does not require position embedding.

\subsubsection{Parallel Structure}
We investigated the performance of EIT$^T$ and Parallel Structure (Fig.\ref{fig:2}-(c)). The results are shown in Table \ref{tab:6}. The accuracy of EIT$^T$ is on average 2.2\% higher than that of Parallel Convolution. Additionally, comparing with Parallel Convolution, EIT has only \textbf{50\%} parameters and \textbf{50\%} FLOPs.
\begin{table}[ht]
\centering
\fontsize{6}{8}\selectfont
\setlength{\tabcolsep}{0.6mm}{
\begin{tabular}{|c|c|c|c|c|c|}
\hline
\multirow{2}{*}{\begin{tabular}[c]{@{}c@{}}TrEn with \\ structure\end{tabular}} & FLOPs & Pram. & Cifar10                   & Cifar100                  & \multirow{2}{*}{Avg.} \\ \cline{2-5}
&\multicolumn{4}{c}{image size:32,32} &                      \\ \hline
EIT$^T$                                                                                             & \textbf{0.428G}       & \textbf{3.095M}                                                   & \textbf{0.855}                          & \textbf{0.605}                          & \textbf{0.730}                     \\ \hline
\begin{tabular}[c]{@{}c@{}}Parallel \\Structure\end{tabular}                                    & 0.887G      & 6.589M                                                   & 0.841                          & 0.574                          & 0.708 \\
\hline
\end{tabular}}
\caption{Ablations on Parallel Convolution.}
\label{tab:6}
\end{table}

\subsubsection{Complicating EIT$^T$}
We try to complicate EIT$^T$ to see if it brings an improvement in performance. For example, we try to add multiple convolutional layers, activation layers, normalization layers, and fully connected layers. The results are shown in Table \ref{tab:7} showing that the one convolutional layer is the best.

\begin{table}[ht]
\centering
\fontsize{6}{8}\selectfont
\setlength{\tabcolsep}{0.6mm}{
\begin{tabular}{|c|c|c|c|c|c|}
\hline
 \multirow{2}{*}{\begin{tabular}[c]{@{}c@{}}Structure\\ of EIT$^T$\end{tabular}}    & FLOPs                   & \begin{tabular}[c]{@{}c@{}}Pram.\end{tabular} & Cifar10           & Cifar100          & \multirow{2}{*}{Avg.} \\ \cline{2-5}
& \multicolumn{4}{c}{Image size:32,32}  &                          \\ \hline
Conv                                                                        & \textbf{0.428G}                  & \textbf{3.095M}                                              & \textbf{0.855}                 & \textbf{0.605}                 & \textbf{0.730}                          \\
 None                                                                        & 0.527G                  & 3.793M                                              & 0.746                 & 0.479                & 0.613                    \\
\begin{tabular}[c]{@{}c@{}} Conv$\times$3\end{tabular} & 0.430G & 3.104M                             & 0.843 & 0.593 & 0.718        \\
\begin{tabular}[c]{@{}c@{}} GELU, Conv, Fc\end{tabular}  & 0.438G       & 3.170M                                   & 0.846 & 0.573 & 0.710        \\
\begin{tabular}[c]{@{}c@{}} Conv, BN, RELU\end{tabular}   & 0.429G       & 3.096M                                  & 0.851 & 0.577 & 0.714  \\
\hline
\end{tabular}}
\caption{Ablations on Complicating EIT$^T$.}
\label{tab:7}
\end{table}

\subsubsection{Increasing and Invariant}
We examine the performance of the Decreasing Structure of EIT$^T$, and the results are shown in Table \ref{tab:8}. Compared with the Invariant Structure and Increasing Structure, the accuracy of Decreasing Structure is 9\% higher on average.

\subsubsection{Removing Position Embedding}
Considering introducing convolutional operations in the network, we investigate whether it still requires position embedding. The results are shown in Table \ref{tab:9} illustrate that the impact of removing position embedding on model performance is negligible. The network without position embedding offers the possibility of simplified adaptation to more visual tasks without the need to redesign embedding.

\begin{table}[h]
\vspace{-0.4cm}
\begin{floatrow}
\capbtabbox{
\centering
\fontsize{6}{8}\selectfont
\setlength{\tabcolsep}{0.6mm}{
\begin{tabular}{|c|c|c|c|c|}
\hline
\multirow{2}{*}{\begin{tabular}[c]{@{}c@{}}Structure\end{tabular}} & Cifar10 & Cifar100                   & \multirow{2}{*}{Avg.}  \\
\cline{2-3}
                      &\multicolumn{2}{c}{Image size:32,32} &                           \\
\hline
Decreasing & \textbf{0.855}       & \textbf{0.605}                          & \textbf{0.730}                          \\
Increasing & 0.790       & 0.481                           & 0.636                          \\
Invariant & 0.817       & 0.476                           & 0.647 \\
\hline
\end{tabular}}
}{
 \caption{Ablations on Increasing and Invariant structure.}
 \label{tab:8}
}

\capbtabbox{
\centering
\fontsize{6}{8}\selectfont
\setlength{\tabcolsep}{0.6mm}{
\begin{tabular}{|c|c|c|c|}
\hline
\multirow{2}{*}{\begin{tabular}[c]{@{}c@{}}Position\\ Embedding\end{tabular}} & Cifar10 & Cifar100     & \multirow{2}{*}{Avg.}  \\
\cline{2-3}
                      &                  \multicolumn{2}{c}{Image size:32,32} &                           \\
\hline
None                        & 0.856          & 0.600                    & 0.728                          \\
Trainable                   & 0.855          & 0.605                    & 0.730\\
\hline
\end{tabular}}
}{
 \caption{Ablations on position embedding.}
 \label{tab:9}
}
\end{floatrow}
\vspace{-0.4cm}
\end{table}

\section{Conclusion}
In this paper, we discuss why the introduction of IB improves ViT's performance. Based on our analysis, we present a simple yet efficient network architecture that introduces IB to ViT with fewer parameters and FLOPs, called EIT. EIT ensures the efficiency of introducing IB without destroying the unification of the network in CV and NLP. Extensive experiments validate that the EIT achieves competitive performance compared with the previous representative ViTs. To the best of our knowledge, we find for the first time a strong correlation between the performance of Transformers and the diversity of head-attention distance, which gives new ideas for further improving the performance of the transformer.





\bibliographystyle{IEEEbib}
\bibliography{icme2022template}

\appendix
\section{Formulism EIT for PaPr}
Formally, given a 2D image $x\in\mathbb{R}^{{H}\times{W}\times{C}}$, we learn a function $f(\cdot)$ that maps $x_0$ into new embeddings $f(x)\in\mathbb{R}^{{H^{'}}\times{W^{'}}\times{C^{'}}}$. $f(\cdot)$ is 2D convolution operation of kernel number $C^{'}$, kernel size $k$, stride $s$ (in ViT, $s$ = $k$, but in this work, $s$ \textless $k$) and $p$ padding. The height and width of the new embedding $f(x)$ take the following values.

\begin{eqnarray}
H^{'} = \lfloor \frac{H + 2p - k}{s} + 1\rfloor, W^{'} = \lfloor \frac{W + 2p - k}{s} + 1\rfloor
\label{eq:6}
\end{eqnarray}where $\lfloor \cdot \rfloor$ denotes rounding down. The height and width of $f(x)$ then reduced by a maximum pooling ($maxpool$) layer with kernel size and stride of $s_m$. By adjusting $s_m$, we can reduce the redundant patches introduced by the convolution operation. $maxpool(f(x))\in\mathbb{R}^{{H_0}\times{W_0}\times{C_0}}$, where $H_0<H^{'},W_0<W^{'},C_0=C^{'}$. Finally, $\mathbb{R}^{{H_0}\times{W_0}\times{C_0}}$ is transformed into $\mathbb{R}^{{H_0}{W_0}\times{C_0}}$ as the final output of EIT$^P$. The above descriptions can be summarized in the following expression.
\begin{eqnarray}
\begin{split}
x_0 = EIT^P(x) = ReShape1D(MaxPool(Conv2D(x)))
\end{split}
\label{eq:7}
\end{eqnarray}

\section{Formulism EIT for TrEn}
Formally, given the normalizated input $x_{i} \in \mathbb{R}^{(1+H_{i}W_{i})\times{C_{i}}}$ ('1' represents the class embedding) of the $i$-layer, different channel dimensions of data are processed by MHA and EIT$^T$, respectively.

\begin{eqnarray}
x_i^{T} \gets EIT^T(x_i[:,~~~:C_i^{T}])
\label{eq:8}\\
x_i^{M} \gets M\!H\!A(x_i[:,-C_i^{M}:])
\label{eq:9}
\end{eqnarray}where $C_i^{T}$ and $C_i^{M}$ are the number of channel dimensions processed by EIT$^T$ and MHA, respectively, satisfying $C_i^{T}+C_i^{M}=C_i$. The $[:]$ is the slice operation. The final output is the combination of the output of the MHA and the EIT$^T$ along the channel dimension.
\begin{eqnarray}
x_i \gets [x_i^{T};x_i^{M}]
\label{eq:10}
\end{eqnarray}where the $[;]$ is the splicing operation of the channel dimension. The MHA is the same operation as ViT \cite{Dosovitskiy2021An}. For each element in the patches $x_{i} \in \mathbb{R}^{(1+H_{i}W_{i})\times{C_{i}^M}}$, we compute a weighted sum over all values $v$ in the patches. The attention weights $A_{mn}$ are based on the pairwise similarity between two elements of the patches and their respective query $q_m$ and key $k_n$ representations.
\begin{eqnarray}
[q, k, v] = x_iU_{qkv}, U_{qkv}\in \mathbb{R}^{C_{i}^M \times C_{i}^{Mh}}
\label{eq:qkv}
\end{eqnarray}
\begin{eqnarray}
A = softmax(qk^\top / \sqrt{C_{i}^{Mh}}), A\in \mathbb{R}^{(1+H_{i}W_{i}) \times (1+H_{i}W_{i})}
\label{eq:A}
\end{eqnarray}
\begin{eqnarray}
HA(x_i) = Av
\label{eq:HA}
\end{eqnarray}
Multi-Head Attention (MHA) is an extension of HA in which we run $k$ attention operations, called "heads", in parallel, and project their concatenated outputs. To keep compute and number of parameters constant when changing $k$, $C_{i}^{Mh}$ is typically set to $C_{i}^{M}/k$.
\begin{eqnarray}
\begin{aligned}
&MHA(x_i) = \\ &[HA_1(x_i[:,:C_{i}^{Mh}]);...;HA_k(x_i[:,(k - 1)C_{i}^{Mh}:])]U_{mha}, \\ &U_{mha}\in \mathbb{R}^{C_{i}^M \times C_{i}^M}
\label{eq:MHA}
\end{aligned}
\end{eqnarray}

Since the convolution in EIT$^T$ handles two-dimension (2D) data, some dimensional transformations are involved before and after the convolution operation. Additionally, EIT$^T$ does not model the class embedding because it is challenging to perform 2D convolution operations if it is added.

\begin{eqnarray}
&EIT^T(x_i) = [x_i^{'};x_i[0,:]]
\label{eq:11}\\
&x_i^{'} \gets Reshape1D(Conv2D(Reshape2D(x_i[1:,:])))
\label{eq:12}
\end{eqnarray}To ensure that $C_i^{M}$ is divisible by $h$ while $C_i^{T}$ decreases layer by layer, for a network with a total of $L$-layer encoders, the $C_i^{T}$ of layer $i$ is set to
\begin{eqnarray}
C_i^{T} = C_i - \lfloor C_i//h \times r_i \rfloor \times h
\label{eq:8}
\end{eqnarray}where $//$ denotes integer division, and $h$ is the number of heads in MHA, which generally requires $C_i$ to divide $h$. $r_i=i/L$, is the division ratio of $C_i^{T}$ to $C_i^{M}$.

\section{Efficiency Considerations}
We utilize standard convolutions in EIT$^P$ and efficient convolutions in EIT$^T$. The convolutions in EIT$^P$ are mainly for high-dimensional mapping, so the standard convolutions with high modelling capability are used. The parameters of EIT$^P$'s convolutions are the same as ViT. Since the strides of EIT$^P$ is smaller than the kernel size, the FLOPs of EIT$^P$ are more than that of ViT. The redundant FLOPs are $\mathcal{O}((\frac{k^2}{s^2}-1)3CHW)$, where $C$ is the channel dimension and 3 is the RGB channels of images. The redundant FLOPs satisfy the linear $C$ complexity and can be neglected compared with the $C^2$ complexity of TrEns. The maximum pooling in EIT$^P$ does not require parameters, and the FLOPs are required to be $\mathcal{O}(HWC)$. It is also a linear $C$ complexity and can be neglected compared with the $C^2$ complexity of TrEns.

Since the convolutions in EIT$^T$ are mainly for increasing the locality of the same channel, the efficient convolutions are used. We split the standard $k\times k$ convolution into a depth-wise separable convolution \cite{howard2017mobilenets} (In order not to slow down the GPU training speed, we just used one of the $depthwise$ operations). Such the convolution requires $k^2C$ parameters and $\mathcal{O}(k^2CT)$ FLOPs, where $T$ is the number of patches for processing. The MHA requires $4C^2+4C$ parameters and $\mathcal{O}(4C^2T+T^2C)$ FLOPs, which means that EIT$^T$ effectively reduces the parameters and FLOPs by replacing the $C^T$ channels of MHA with the depth-wise separable convolution.

\end{document}